\documentclass[runningheads]{llncs}
\usepackage{makeidx}
\usepackage{graphicx}
\usepackage{amsmath,amssymb} 
\usepackage{color}

\usepackage{times}
\usepackage{epsfig}
\usepackage{graphicx}
\usepackage{amsmath}
\usepackage{amssymb}
\usepackage[utf8]{inputenc} 

\usepackage{tabularx}
\usepackage{multirow} 
\usepackage{booktabs} 
\usepackage{adjustbox}

\usepackage{pgfplots}          
\usepgfplotslibrary{groupplots}
\pgfplotsset{compat=1.13} 

\usepackage{tikz}
\usetikzlibrary{calc} 
\usetikzlibrary{arrows} 
\usetikzlibrary{positioning} 
\usetikzlibrary{fit} 

\usepackage{bm} 
\usepackage{subcaption}
\usepackage{rotating}

\usepackage{siunitx} 
\usepackage{float}
\usepackage{mycommands}
\usepackage{nicefrac}

\newcolumntype{L}[1]{>{\raggedright\let\newline\\\arraybackslash\hspace{0pt}}m{#1}}
\newcolumntype{C}[1]{>{\centering\let\newline\\\arraybackslash\hspace{0pt}}m{#1}}
\newcolumntype{R}[1]{>{\raggedleft\let\newline\\\arraybackslash\hspace{0pt}}m{#1}}
\newcolumntype{M}[1]{>{\centering\arraybackslash}m{#1}}
\newcolumntype{N}{@{}m{0pt}@{}}

\usepackage{rotating}
\usepackage{array,makecell,multirow}

\usepackage[capitalise]{cleveref}

\graphicspath{{figures/}}

\begin{document}
	\pagestyle{headings}
	\mainmatter

	\def\GCPR19SubNumber{41}

	\title{Learned Collaborative Stereo Refinement}

	\titlerunning{Learned Collaborative Stereo Refinement}
	\authorrunning{P. Knöbelreiter and T. Pock}
	\author{Patrick Knöbelreiter\inst{1}\orcidID{0000-0002-2371-014X} and Thomas Pock\inst{1}\orcidID{0000-0001-6120-1058}\\
	        \texttt{\{knoebelreiter, pock\}@icg.tugraz.at}}
	\institute{$^1$Institute for Computer Graphics and Vision, Graz University of Technology}

		

	\maketitle

	\begin{abstract}
		In this work, we propose a learning-based method to denoise  and refine disparity maps of a given stereo method. The proposed variational network arises naturally from unrolling the iterates of a proximal gradient method applied to a variational energy defined in a joint disparity, color, and confidence image space. Our method allows to learn a robust collaborative regularizer leveraging the joint statistics of the color image, the confidence map and the disparity map. Due to the variational structure of our method, the individual steps can be easily visualized, thus enabling interpretability of the method. We can therefore provide interesting insights into how our method refines and denoises disparity maps. The efficiency of our method is demonstrated by the publicly available stereo benchmarks Middlebury 2014 and Kitti 2015.

 \end{abstract}
 
 \section{Introduction}
\label{sec:intro}

Computing 3D information from a stereo pair is one of the most important problems in computer vision.
One reason for this is that depth information is a very strong cue to understanding visual scenes, and depth information is therefore an integral part of many vision based systems.
For example, in autonomous driving, it is not sufficient to know the objects visible in the scene, but it is also important to estimate the distance to these objects. 
A lidar scanner is often too expensive and provides only sparse depth estimates.
Therefore, it is an interesting alternative to compute depth information exclusively from stereo images. 
However, the calculation of depth information from images is still a very challenging task.
Reflections, occlusions, challenging illuminations \etc. make the task even harder.
To tackle these difficulties the computation of dense depth maps is usually split up into the four steps (i) matching cost computation, (ii) cost aggregation, (iii) disparity computation and (iv) disparity refinement \cite{Scharstein2002}.
In deep learning based approaches (i) and (ii) are usually implemented in a matching convolutional neural network (CNN), (iii) is done using graphical models or 3D regularization CNNs and (iv) is done with a refinement module \cite{Tulyakov2018_NIPS}. 

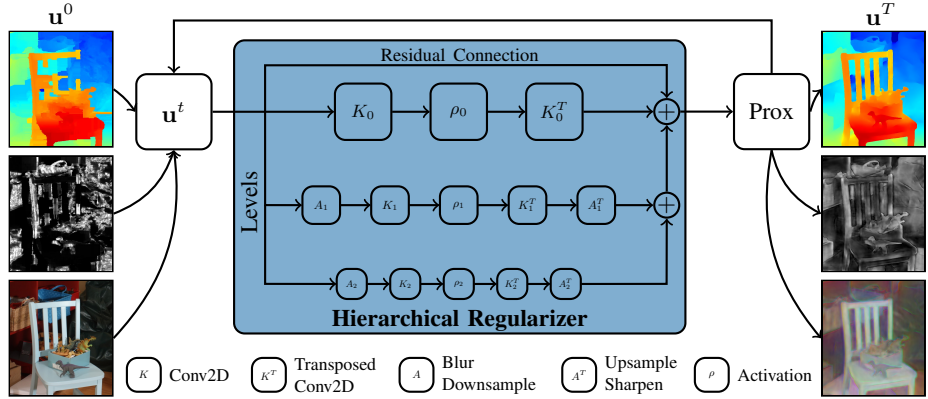
\begin{figure}
  \begin{tikzpicture}
    \begin{scope}[thick]
    \begin{scope}[minimum height=1cm, minimum width=1cm, align=center, rounded corners, inner sep=0, node distance=0.5cm, anchor=center]
      \node[draw, ultra thick, minimum width=1.2cm, rounded corners=0] (rgb) 
      {\includegraphics[height=1.5cm, clip, trim={300 0 650 450}]{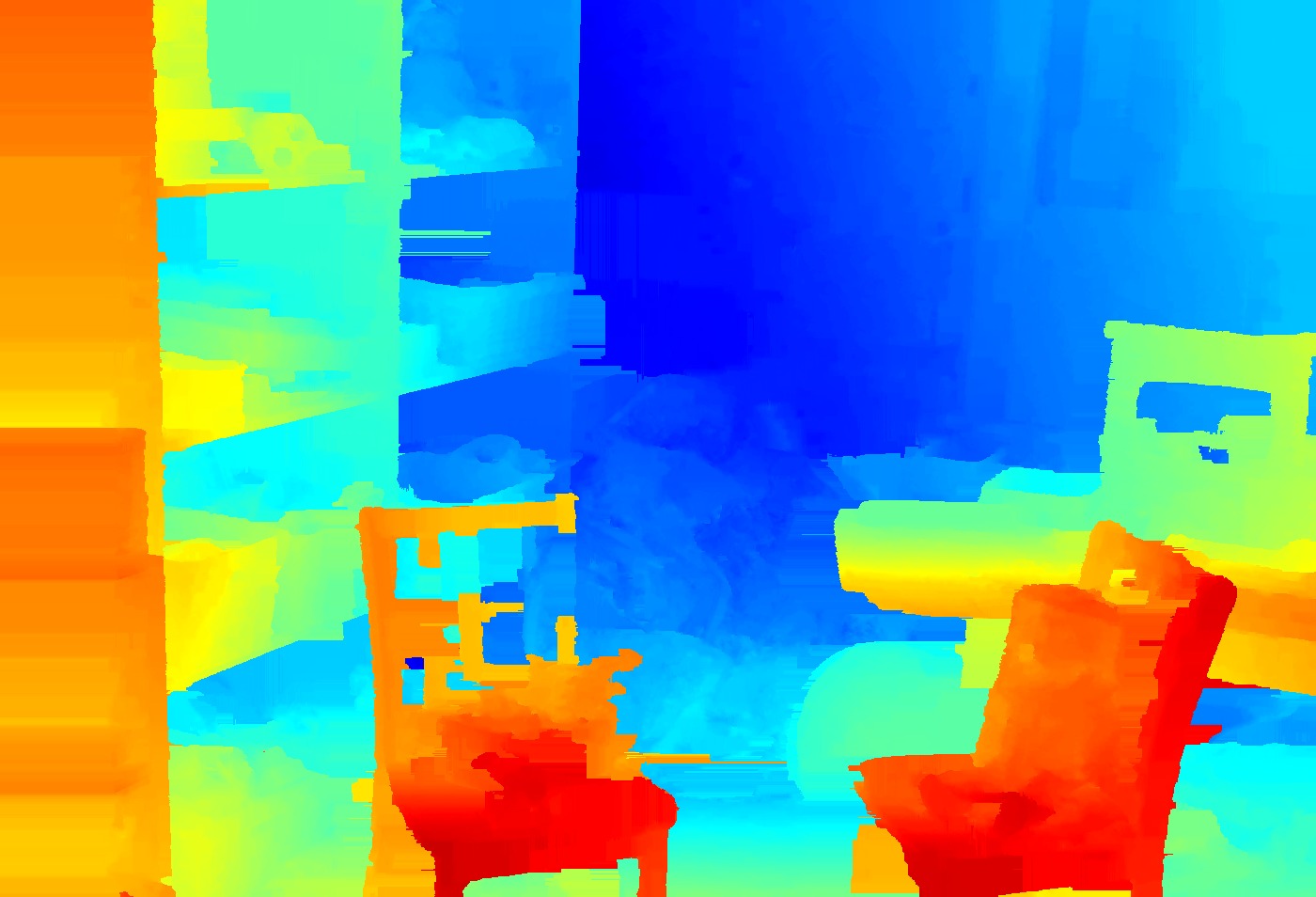}};
      \node[draw, ultra thick, minimum width=1.2cm, rounded corners=0, below=0.1cm of rgb] (conf) 
      {\includegraphics[height=1.5cm, clip, trim={195 0 650 450}]{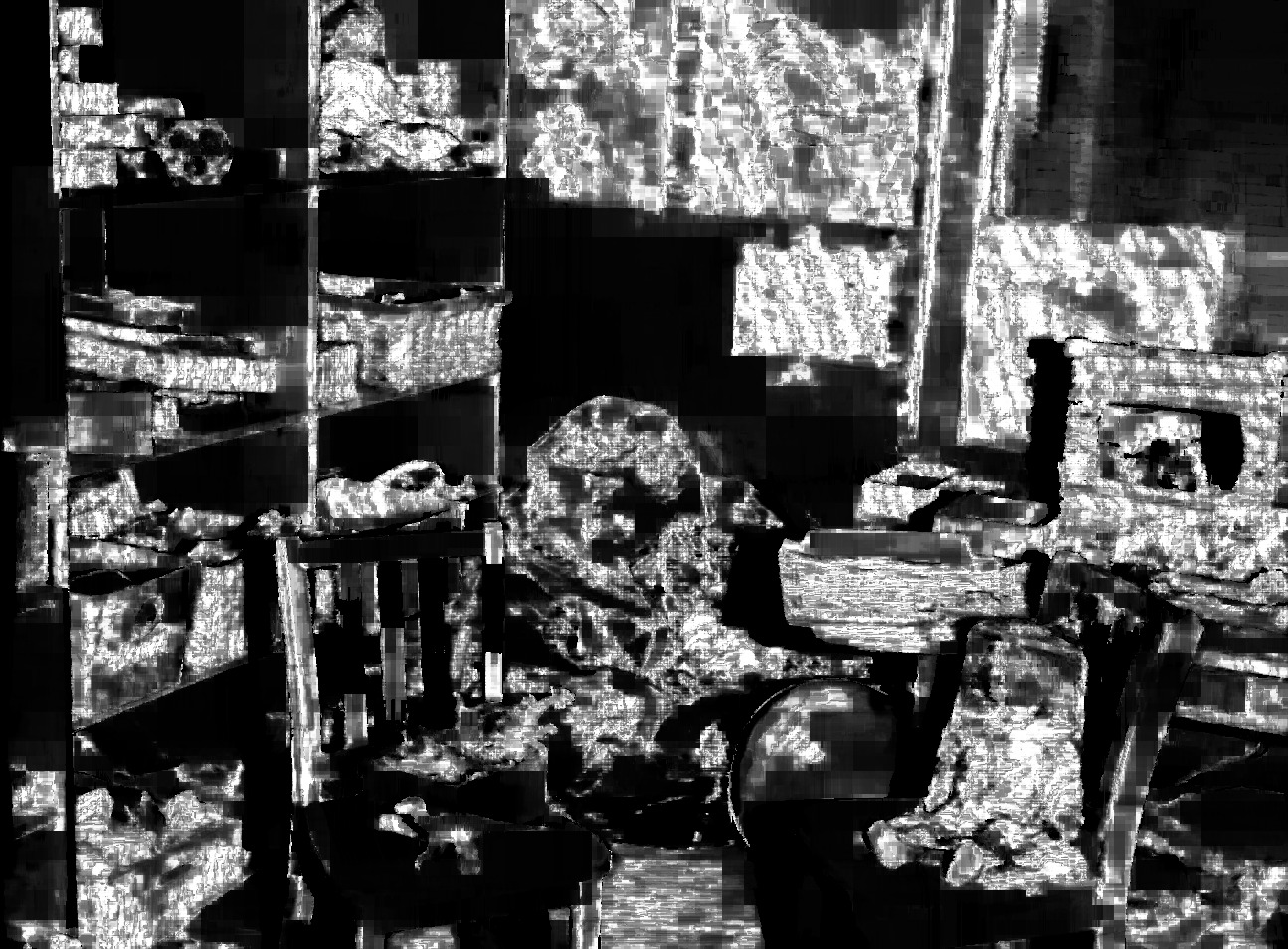}};
      \node[draw, ultra thick, minimum width=1.2cm, rounded corners=0, below=0.1cm of conf] (disp) 
      {\includegraphics[height=1.5cm, clip, trim={300 0 650 450}]{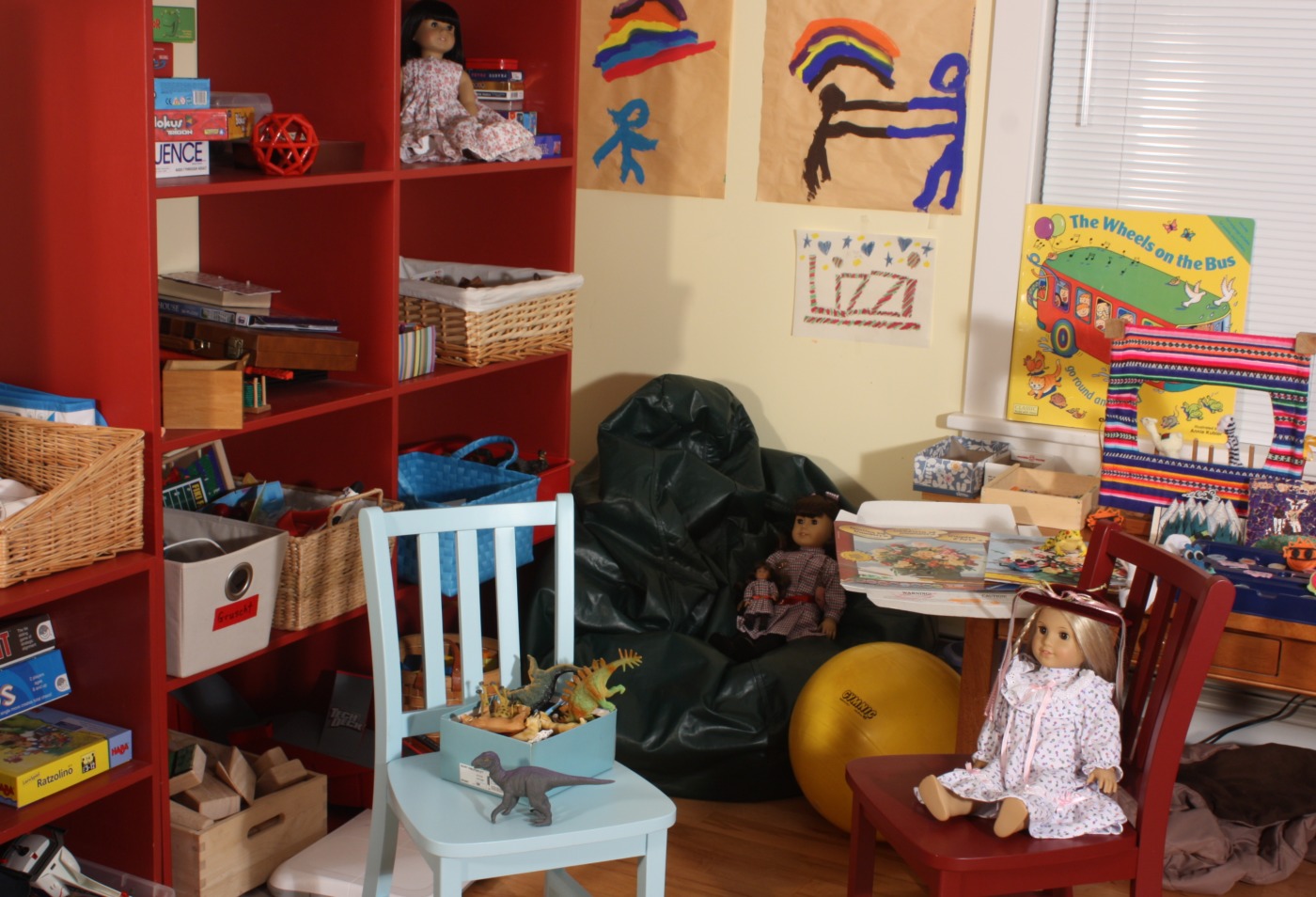}};

      \node[draw, minimum width=6cm, minimum height=3.9cm, fill={rgb:red,0;green,108;blue,186}, fill opacity=0.5] (reg) at (5.3, -1.3) {};
      \node[below=-0.7cm of reg,text width=5cm] (regcaption) {\textbf{Hierarchical Regularizer}};
      \node[above=-0.725cm of reg,text width=5cm] (residual) {{\scriptsize Residual Connection}};

      \node[] at (0, 1) {$\mathbf{u}^0$};
      \node[] at (10.9, 1) {$\mathbf{u}^T$};

      \node[draw] (input) at (1.5, -0.3) {$\mathbf{u}^t$};

      \draw[->] (rgb.east) to [bend left=10] (input.west);
      \draw[->] (conf.east) to [bend right=25] (input.south);
      \draw[->] (disp.east) to [bend right=25] (input.south);
    
      \begin{scope}[every node/.style={scale=0.75}]
      \node[right=0.5cm of input, minimum size=0.5cm] (scales) {};
      \node[above=0.57cm of scales.center, minimum size=0.1cm] (scalesdummy) {};
      \node[draw, right=0.72cm of scales] (K0) {$K_0$};
      \draw[->] (input) -- (K0);
      \node[draw, right=of K0] (phi0) {$\rho_0$};
      \draw[->] (K0) -- (phi0);
      \node[draw, right=of phi0] (K0T) {$K_0^T$};
      \draw[->] (phi0) -- (K0T);

      \begin{scope}[every node/.style={scale=0.5}]
        \node[draw, below=0.6cm of phi0] (phi1) {$\rho_1$};
        \node[draw, left=0.4cm of phi1] (K1) {$K_1$};
        
        \draw[->] (K1) -- (phi1);
        \node[draw, right=0.4cm of phi1] (K1T) {$K_1^T$};
        \draw[->] (phi1) -- (K1T);
        \node[draw, left=0.4cm of K1] (A1) {$A_1$};
        \node[draw, right=0.4cm of K1T] (A1T) {$A_1^T$};
        \draw[->] (A1) --(K1);
        \draw[->] (K1T) -- (A1T);
      \end{scope}
      \begin{scope}[every node/.style={scale=0.4}]
        \node[draw, below=1.7cm of phi0] (phi2) {$\rho_2$};
        \node[draw, left=0.3cm of phi2] (K2) {$K_2$};
        \draw[->] (K2) -- (phi2);
        \node[draw, right=0.3cm of phi2] (K2T) {$K_2^T$};
        \draw[->] (phi2) -- (K2T);

        \node[draw, left=0.3cm of K2] (A2) {$A_2$};
        \draw[->] (A2) -- (K2);

        \node[draw, right=0.3cm of K2T] (A2T) {$A_2^T$};
      \end{scope}
    \end{scope}
  \end{scope}

    \draw[->] (scales.center) -- (2.7125, -1.542) -- (A1.west);
    \draw[->] (2.7125, -1.542) -- (2.7125, -2.59) -- (A2.west);
    
    \node[draw, circle, right=0.5cm of A1T, minimum size=0cm, inner sep=0pt] (p1) {$+$};
    \node[draw, circle, above=0.8845cm of p1, minimum size=0cm, inner sep=0pt] (p0) {$+$};
    \node[above=0.5cm of p0.center, minimum size=0.1cm] (p0dummy) {};

    \begin{scope}[minimum height=1cm, minimum width=1cm, align=center, rounded corners, inner sep=0]
      \draw[->] (K0T) -- (p0); 

      \draw[->, rounded corners=0] (scales.center) --(scalesdummy.center) -- (p0dummy.center) -- (p0.north);

      \node[draw, right=0.7cm of p0] (prox) {Prox};
      \draw[->] (p0) to  (prox);
      \draw[->] (p1) -- (p0);
      \draw[->] (K2T) -- (A2T);
      \draw[->] (A1T) -- (p1);

      \draw[->, rounded corners=0cm] (A2T.east) -- (8.05, -2.59) -- (p1.south);

      \node[draw, ultra thick, minimum width=1.2cm, rounded corners=0] (rgbT) at (10.8, 0) 
      {\includegraphics[height=1.5cm, clip, trim={300 0 650 450}]{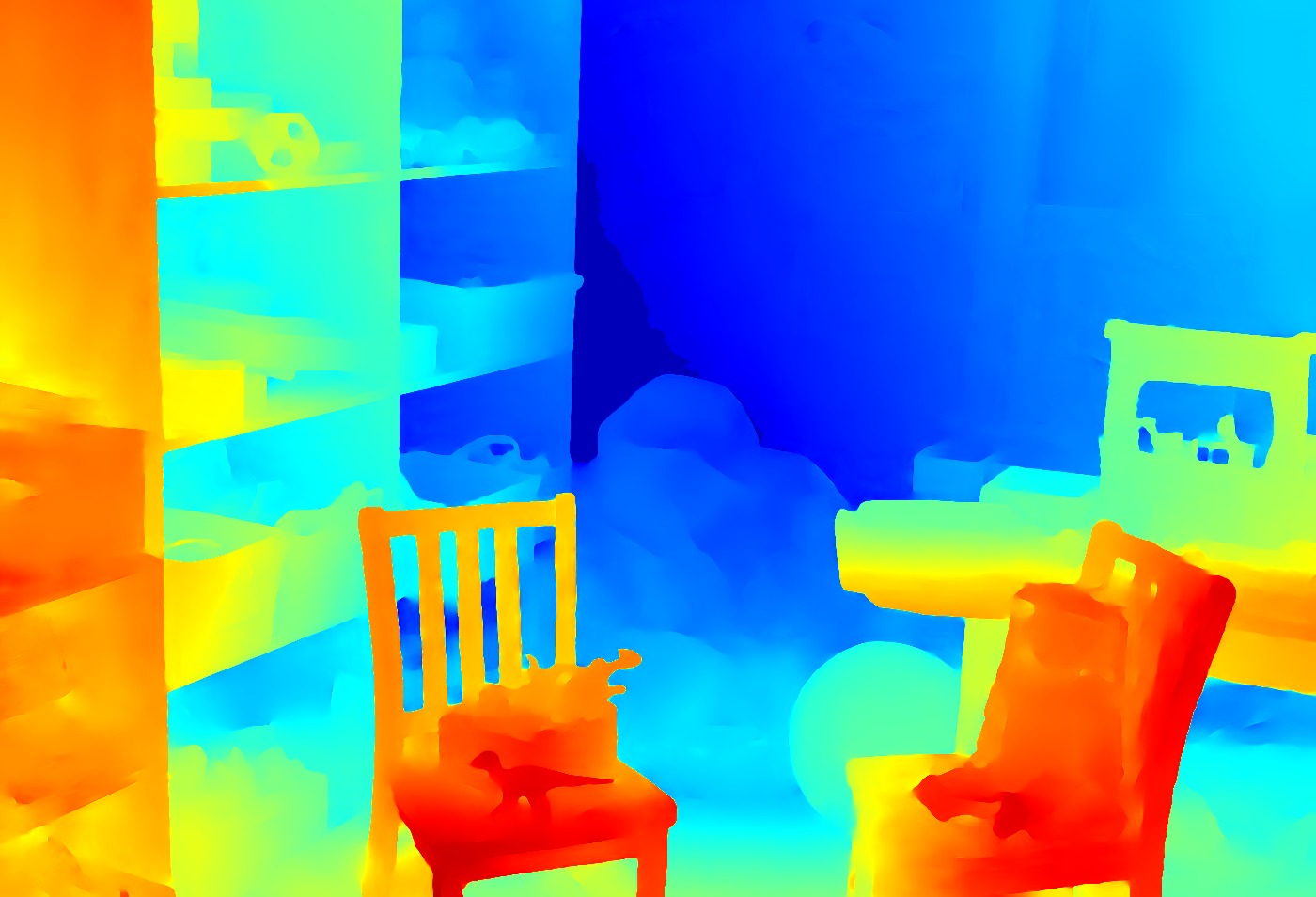}};
      \node[draw, ultra thick, minimum width=1.2cm, rounded corners=0, below= 0.1cm of rgbT] (confT){\includegraphics[height=1.5cm, clip, trim={300 0 650 450}]{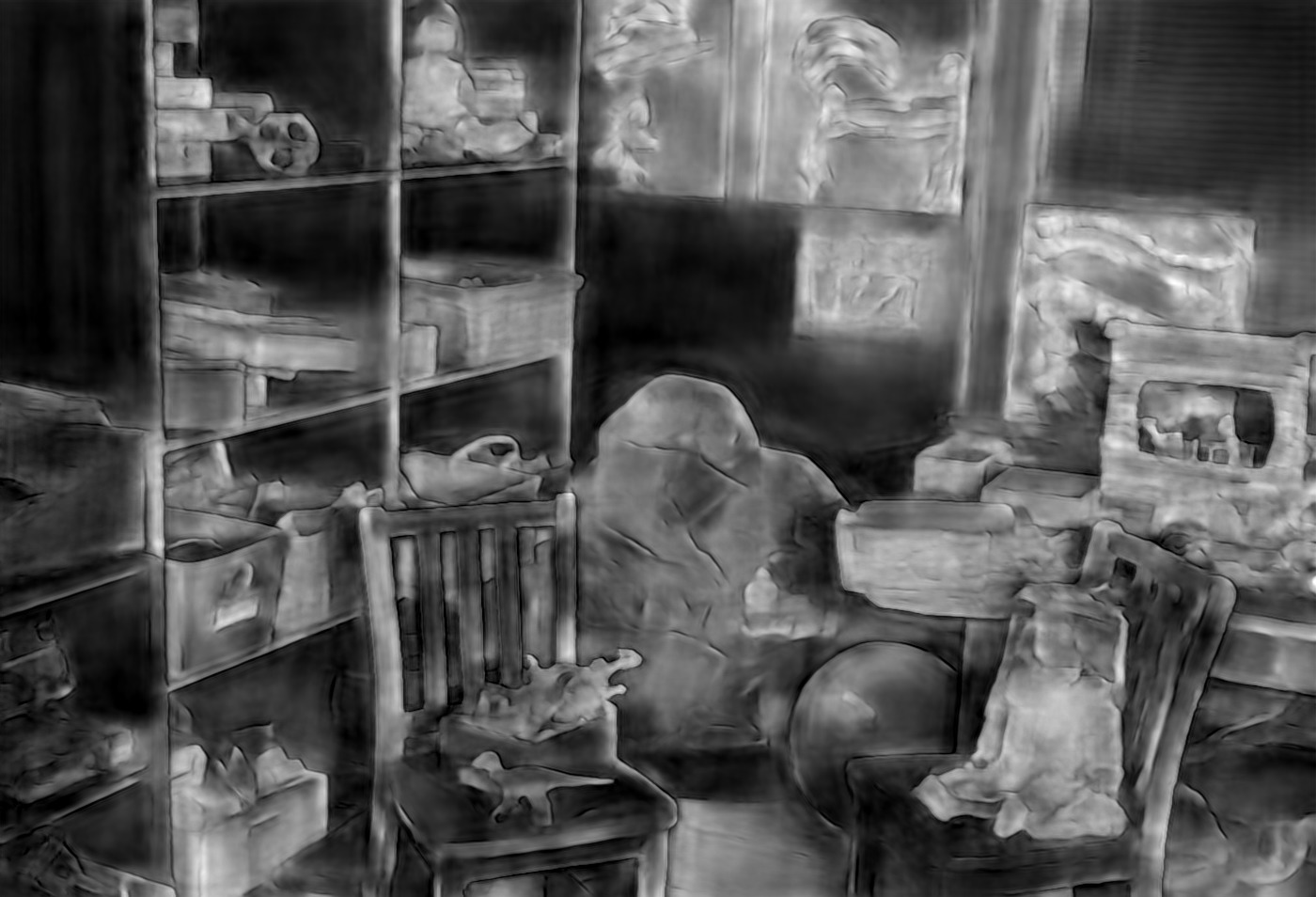}}; 
      \node[draw, ultra thick, minimum width=1.2cm, rounded corners=0, below=0.1cm of confT] (dispT)
      {\includegraphics[height=1.5cm, clip, trim={300 0 650 450}]{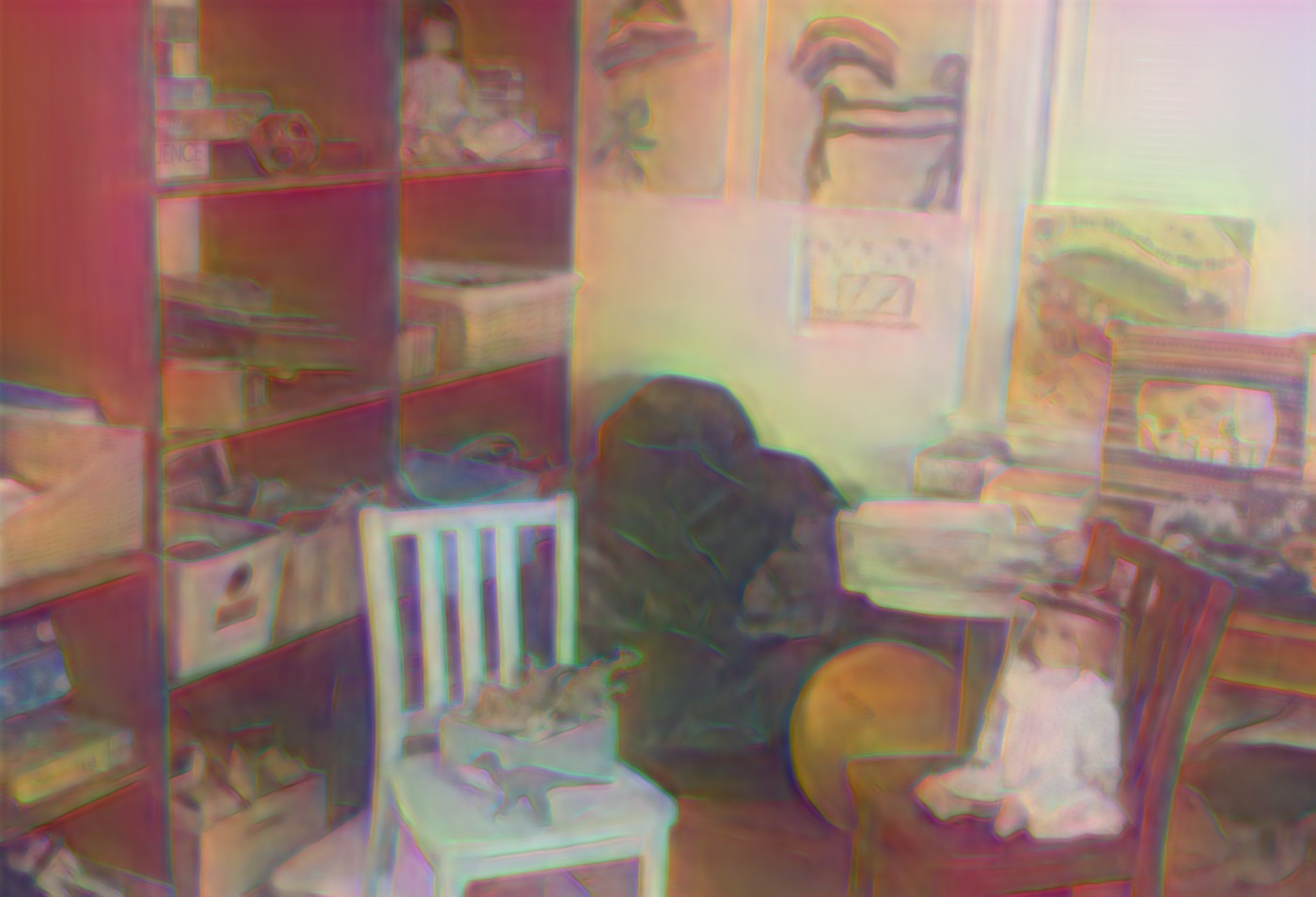}};

      \node[above=0.7cm of prox] (dummy) {};
      \node[above=0.7cm of input] (dummy1) {}; 

      \draw[->, rounded corners=0] (prox) -- (dummy.south) -- (dummy1.south) -- (input);

      \draw[->] (prox.east) to [bend left=10] (rgbT.west);
      \draw[->] (prox.south) to [bend right=25] (confT.west);
      \draw[->] (prox.south) to [bend right=25] (dispT.west);

      \node[rotate=90] () at (2.5, -1.5) {Levels};

      \begin{scope}[every node/.style={scale=0.45}]
        \node[draw] (legendK) at(1.1, -3.8) {$K$};
        \node[draw, right=1.2cm of legendK] (legendKT) {$K^T$};
        \node[draw, right=1.5cm of legendKT] (A) {$A$};
        \node[draw, right=1.7cm of A] (AT) {$A^T$};
        \node[draw, right=1.3cm of AT] (acti) {$\rho$};
      \end{scope}
      \begin{scope}[every node/.style={scale=0.75}, align=left]
        \node[right=0.1cm of legendK] () {Conv2D}; 
        \node[right=0.1cm of legendKT] () {Transposed\\Conv2D};           
        \node[right=0.1cm of A] () {Blur\\Downsample};           
        \node[right=0.1cm of AT] () {Upsample\\Sharpen};           
        \node[right=0.1cm of acti] () {Activation};           
      \end{scope}
      
    \end{scope}

  \end{scope}

  \end{tikzpicture}
  \caption{Model Overview. Our model takes three inputs, an initial disparity map, confidence map and the color image. The collaborative hierarchical regularizer iteratively computes a refined disparity map and yields refined confidences and an abstracted color image as a byproduct. The subscripts indicate the level.}
  \label{fig:overview}
\end{figure}

There are many approaches to tackle (i)-(iii).
However, there are only a few learning-based works for disparity refinement (iv) (see \cref{sec:related}).
Existing work to refine the disparity map is often based on another CNN using residual connections.
In this work we want to overcome these black-box refinement networks with a simple, effective and most important easily interpretable refinement approach for disparity maps.
We tackle the refinement problem with a learnable hierarchical variational network.
This allows us to exploit both the power of deep learning and the interpretability of variational methods.
In order to show the effectiveness of the proposed refinement module, we conduct experiments on directly refining/denoising winner-takes-all (WTA) solutions of feature matching and as a pure post-processing module on top of an existing stereo method.
\cref{fig:overview} shows an overview of our method.
Starting from an initial disparity map the final result is iteratively reconstructed.

\paragraph{Contributions}
We propose a learnable variational refinement network which takes advantage of the joint information of the color image, the disparity map and a confidence map to compute a regularized disparity map. 
We therefore show how our proposed method can be derived from the iterates of a proximal gradient method and how it can be specifically designed for stereo refinement. 
Additionally, we evaluate a broad range of possible architectural choices in an ablation study. 
Furthermore, we give insights into how our model constructs the final disparity map by visualizing and interpreting the intermediate iterates.
We show the effectiveness of our method by participating on the two complementary public available benchmarks Middlebury 2014 and Kitti 2015.

 \section{Related Work}
\label{sec:related}
We propose a learnable variational model, where 
we use the modeling power of variational calculus to explicitly guide the refinement process for stereo.
Thus, we focus on disparity refinement in the following sections.

\paragraph{Variational Methods}
Variational methods formulate the correspondence problem as minimization of an energy functional comprising a data fidelity term and a smoothness term. 
We also briefly review variational optical flow methods, because stereo is a special case of optical flow, \ie~it can be considered as optical flow in horizontal direction only.
The data-term measures usually the raw intensity difference \cite{Brox2004_ECCV,zach2007duality,chambolle2011first} between the reference view and the warped other view. 
The regularizer imposes prior knowledge on the resulting disparity map.
This is, the disparity map is assumed to be piecewise smooth.
Prominent regularizers are the robust Total Variation (TV) \cite{zach2007duality} and the higher order generalization of TV as e.g. used by Ranftl \etal~\cite{Ranftl2012_IVS,ranftl2014non} or by Kuschk and Cremers~\cite{kuschk2013fast}.
Variational approaches have two important advantages in the context of stereo. They  naturally produce sub-pixel accurate disparities and they are easily interpretable.
However, in order to also capture large displacements a coarse-to-fine warping scheme \cite{Brox2004_ECCV} is necessary. 
To overcome the warping scheme without losing fine details, variational methods can also be used to refine an initial disparity map.
This has \eg~be done by Shekhovtsov \etal~\cite{shekhovtsov2016solving} who refined the initial disparity estimates coming from a Conditional Random Field (CRF).
Similarly, \cite{RevaudWHS15} and \cite{Maurer17} used a variational method for refining optical flow.

\paragraph{Disparity Refinement}
Here we want to focus on the refinement of an initial disparity map. 
The initial disparity map can be \eg~the WTA solution of a matching volume or any other output of a stereo algorithm. 
One important approach of refinement algorithms is the fast bilateral solver (FBS) \cite{Barron2016}.
This algorithm refines the initial disparity estimate by solving an optimization problem containing an $\ell_2$ smoothness- and an $\ell_2$ data-fidelity term.
The fast bilateral solver is the most related work to ours. 
However, in this work we replace the $\ell_2$ norm with the robust $\ell_1$ norm.
More importantly, we additionally replace the hand-crafted smoothness term by a learnable multi-scale regularizer.
Another refinement method was proposed by Gidaris and Komodakis~\cite{gidaris2017detect}. 
They also start with an initial disparity map, detect erroneous regions and then replace and refine these regions to get a high-quality output.
Pang \etal~\cite{Pang_2017_ICCV_Workshops} proposed to apply one and the same network twice. 
They compute the initial disparity map in a first pass, warp the second view with the initial disparity map and then compute only the residual to obtain a high quality disparity map.
Liang \etal~\cite{Liang_2018_CVPR} also improved the results by adding a refinement sub-network on top of the regularization network. 
We want to stress that the CNN based refinement networks~\cite{Liang_2018_CVPR,Pang_2017_ICCV_Workshops} do not have a specialized architecture for refinement as opposed to the proposed model.

\paragraph{Learnable optimization schemes}
Learnable optimization schemes are based on unrolling the iterates of optimization algorithms. 
We divide the approaches into two categories. 
In the first category the optimization iterates are mainly used to utilize the structure during learning. 
For example in \cite{riegler2016atgv} 10 iterations of a TGV regularized variational method are unrolled and used for depth super-resolution.  
However, they kept the algorithm fixed, \ie~the only learnable parameters in the inference part are the step-sizes.
Similarly, in \cite{vogel2018learning} unrolling 10k iterations of the FISTA \cite{beck2009fast} algorithm is proposed. 
The second category includes methods where the optimization scheme is not only used to provide the structure, but it is also generalized by adding additional learnable parameters directly to the optimization iterates.
For example \cite{vogel2017primal} proposed a primal-dual-network for low-level vision problems, 
where the authors learned the inference part of a Markov Random Field (MRF) model generalizing a primal-dual algorithm. 
Chen \etal~\cite{chen2015learning} generalized a reaction-diffusion model and successfully learned a model for image denoising. 
Based on~\cite{chen2015learning} a generalized incremental proximal gradient method was proposed in~\cite{Kobler2017_GCPR}, where the authors showed connections to residual units~\cite{He_2016_CVPR}.
We built on the work of Chen \etal, but specially designed the energy terms for the stereo task.
Additionally, we allow to regularize on multiple spatial resolutions jointly and make use of the robust $\ell_1$ function in our data-terms.

 \section{Method}
\label{sec:method}

We consider images to be functions $f: \Omega \rightarrow \mathbb{R}^C$, with $\Omega \subset \mathbb{N}^2_+$ and $C$ is the number of channels which is 3 for RGB color images.
Given two images $f^0$ and $f^1$ from a rectified stereo pair, we want to compute dense disparities $d$ such that
 $f^0(x) = f^1 (x - \tilde d )$,
\ie~we want to compute the horizontal shift $\tilde d = (d, 0)$ for each pixel $x = (x_1, x_2)$ between the reference image $f^0$ and the second image $f^1$.
Here, we propose a novel variational refinement network for stereo which operates solely in 2D image space and is thus very efficient.
The input to our method is an initial disparity map $\check u : \Omega \rightarrow [0, D]$, where $D$ is the maximal disparity, a reference image $f^0$ and a pixel-wise confidence map $c:\Omega \rightarrow [0,1]$. 
The proposed variational network is a method to regularize, denoise and refine a noisy disparity map with learnable filters and learnable potential functions.
Hence, the task we want to solve is the following: Given a noisy disparity map $\check u$, we want to recover the clean disparity with $T$ learnable variational network steps.
We do not make any assumptions on the quality of the initial disparity map, i.e. the initial disparity map may contain many strong outliers.

\subsection{Collaborative Disparity Denoising}
\label{ssec:vn}
\begin{figure}[t]
  \centering
  \begin{subfigure}{0.45\textwidth}
  \includegraphics[width=\textwidth, clip, trim={0 30 0 0}]{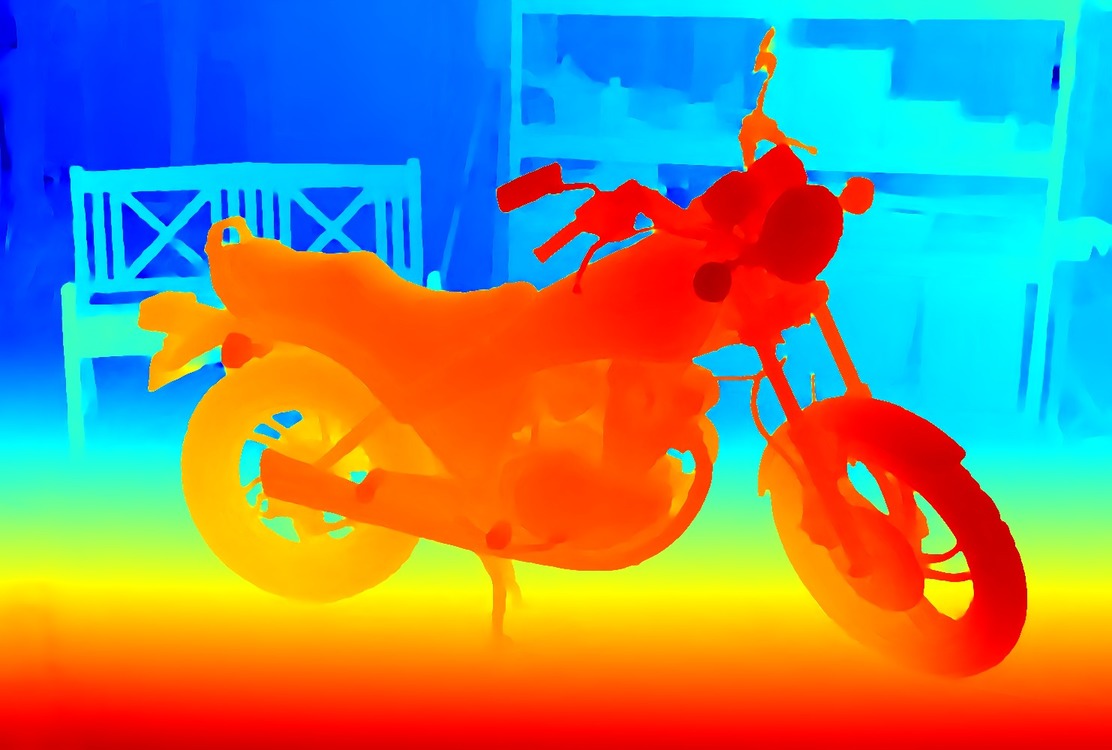}
  \vspace{-0.55cm}
  \caption{VN Disparity Map}  
  \end{subfigure}
  \begin{subfigure}{0.45\textwidth}
  \includegraphics[width=\textwidth, clip, trim={0 30 0 0}]{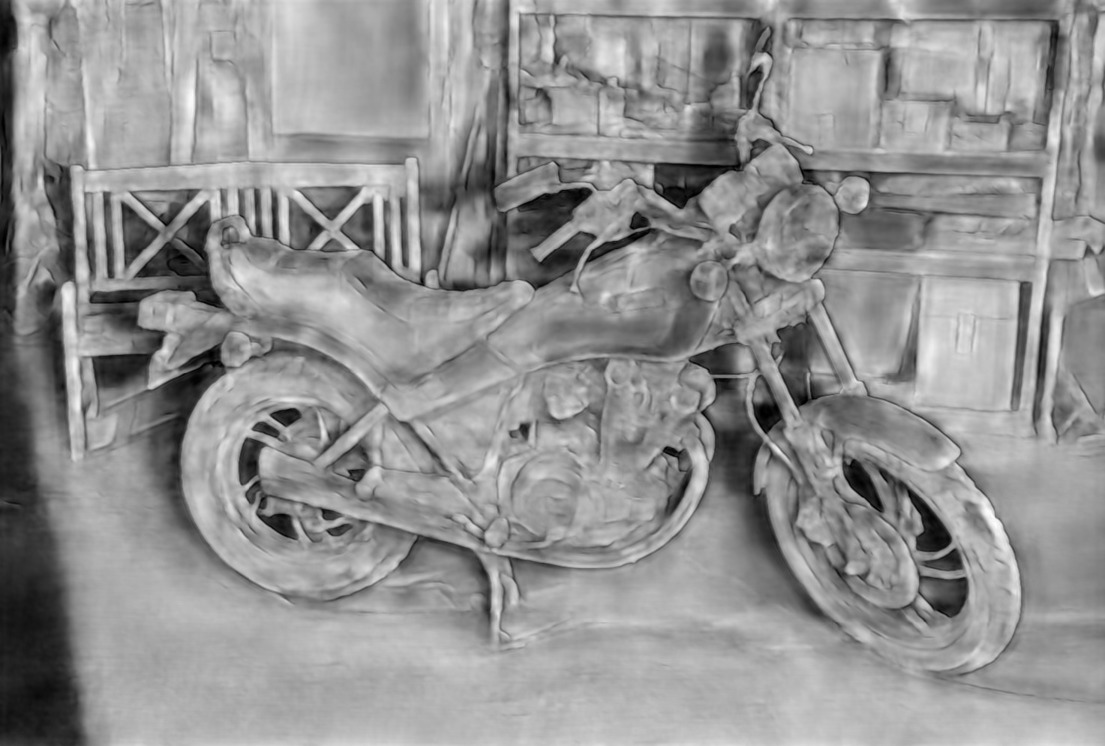}
  \vspace{-0.55cm}
  \caption{VN Confidence Map}
  \end{subfigure}
  \begin{subfigure}{0.45\textwidth}
  \includegraphics[width=\textwidth, clip, trim={0 5 0 0}]{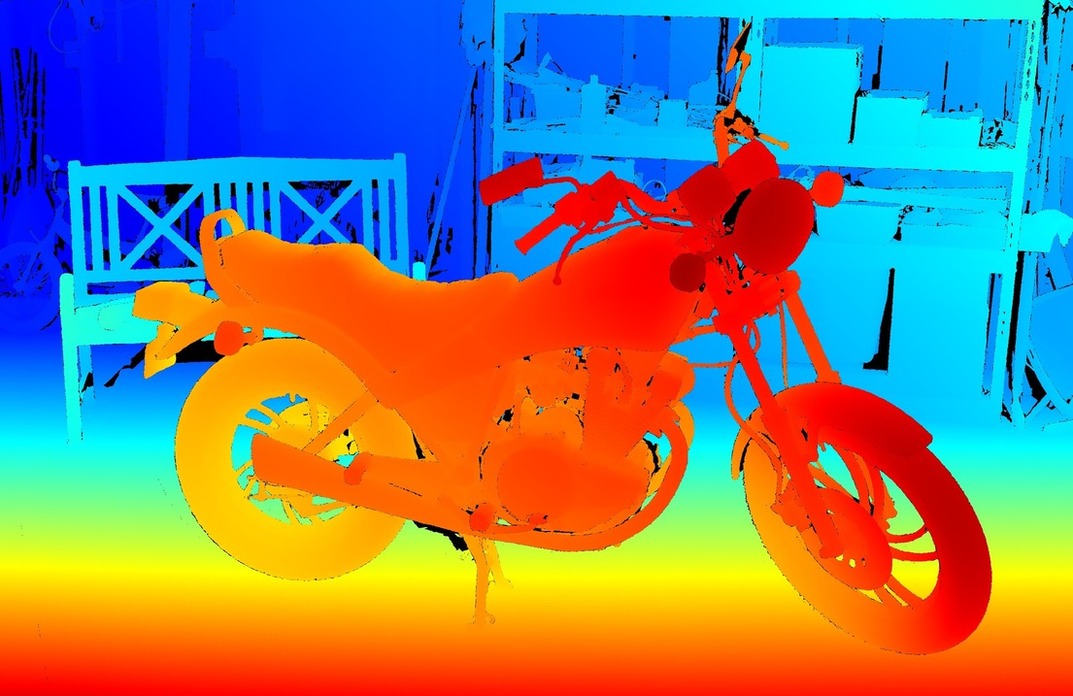}
  \vspace{-0.55cm}
  \caption{Ground-truth}
  \end{subfigure}
  \begin{subfigure}{0.45\textwidth}
  \includegraphics[width=\textwidth, clip, trim={0 30 0 0}]{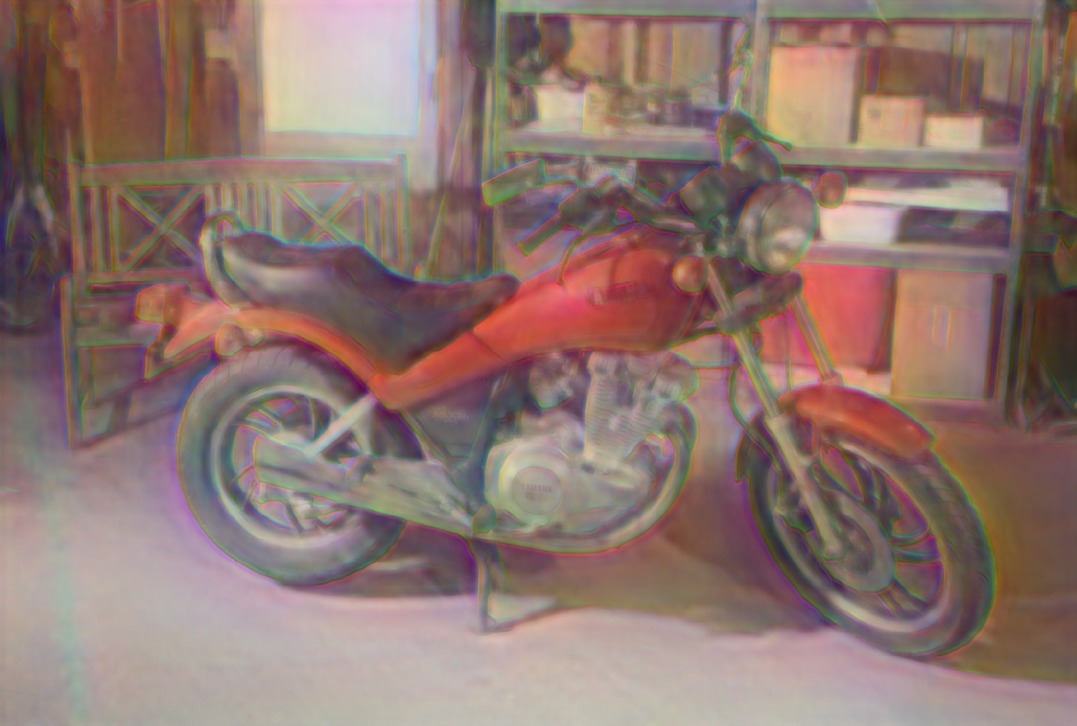}
  \vspace{-0.55cm}
  \caption{VN Color Image}
  \end{subfigure}
  \caption{Collaborative Disparity Denoising. Our method produces three outputs: (a) the refined disparity map, (b) the refined confidence map and (d) the refined color image. (c) shows the ground-truth image for comparison (black pixels = invalid). Note how our method is able to preserve fine details such as the spokes of the motorcycle.}
  \label{fig:initialVsFinal}
\end{figure}
As the main contribution of this paper, we propose a method that performs a collaborative denoising in the joint color image, disparity and confidence space (see \cref{fig:initialVsFinal}).
Our model is based on the following three observations:
(i) Depth discontinuities co-inside with object boundaries, (ii) discontinuities in the confidence image are expected to be close to left-sided object boundaries and (iii) the confidence image can be used as a pixel-wise weighting factor in the data fidelity term.
Based on these three observations, we propose the following collaborative variational denoising model

\begin{equation}
  \min_{\mathbf{u}} \mathcal{R}(\mathbf{u}) 
  +
  \mathcal{D}(\mathbf{u}),
  \label{eq:vnAllgemein}
\end{equation}
where $\mathbf{u} = (\mathbf{u}^{rgb},u^d, u^c) : \Omega \rightarrow \mathbb{R}^5$, \ie~$\mathbf{u}$ contains for every pixel an RGB color information, a disparity value and a confidence value.
$\mathcal{R}(\mathbf{u})$ denotes the collaborative regularizer and it is given by a multi-scale and multi-channel version of the Fields of Experts (FoE) model \cite{Roth2009_FoE} with $L$ scales and $K$ channels.
\begin{equation}
  \mathcal{R}(\mathbf{u}; \theta)= \sum_{l=1}^L \sum_{k=1}^K \sum_{x \in \Omega} \phi_k^l \left( \left( K_k^l A^l \mathbf{u} \right)(x) \right),
\end{equation}
where $A^l$ are combined blur and downsampling operators, $K^l_k$ are linear convolution operators and $\phi^l_k : \mathbb{R} \mapsto \mathbb{R}$ are non-linear activation functions. 
The vector $\theta$ holds the parameters of the regularizer which will be detailed later.
Note that multiple levels allow the model to operate on different spatial resolutions and therefore enables the denoising of large corrupted areas.
Intuitively, the collaborative regularizer captures the statistics of the joint color, confidence and disparity space.
Hence, it will be necessary to learn the linear operators and the non-linear potential functions from data.
It will turn out that the combination of filtering in the joint color-disparity-confidence space at multiple hierarchical pyramid levels and specifically learned channel-wise potential functions make our model very powerful.

$\mathcal{D}(\mathbf{u})$ denotes the collaborative data fidelity term and it is defined by
\begin{equation}
  \mathcal{D}(\mathbf{u}; \theta) = \frac{\lambda}{2}\lVert \mathbf{u}^{rgb} \!\!- \mathbf{f}^0 \rVert^2 + \mu \lVert u^c - c \rVert_1 + \nu \lVert u^d - \check d \rVert_{u^c,1},
\end{equation}
where $\theta$ is again a placeholder for the learnable parameters.
The first term ensures that the smoothed color image $\mathbf{u}^{rgb}$ does not deviate too much from the original color image $\mathbf{f}^0$, the second term ensures that the smoothed confidence map stays close to the original confidence map.
Here we use an $\ell_1$ norm in order to deal with outliers in the initial confidence map.
The last term is the data fidelity term of the disparity map. 
It is given by an $\ell_1$ norm which is pixel-wise weighted by the confidence measure $u^c$.
Hence, data fidelity is enforced in high-confidence regions and suppressed in low-confidence regions.
Note that the weighted $\ell_1$ norm additionally ties the disparity map with the confidence map during the steps of the variational network.

\paragraph{Proximal Gradient Method (PGM)}
We consider a PGM \cite{Neal_FTO_2014} whose iterates are given by
\begin{equation}
  \mathbf{u}_{t+1} = \text{prox}_{\alpha_t \mathcal{D}}(\mathbf{u}_t - \alpha_t \nabla \mathcal{R}(\mathbf{u}_t)), 
  \label{eq:PGM}
\end{equation}
where $\alpha_t$ is the step-size, $\nabla \mathcal{R}(\mathbf{u}_t)$ is the gradient of the regularizer which is given by
\begin{equation}
    \nabla \mathcal{R}(\mathbf{u}) = \sum_{l=1}^{L} \sum_{k=1}^K (K^l_k A^l)^T \rho_k^l \left( K_k^l A^l \mathbf{u} \right),
  \label{eq:grad_step_regularizer}
\end{equation}
where $\rho_k^l = \mathrm{diag}((\phi_k^l)')$.
$\text{prox}_{\alpha_t \mathcal{D}}$ denotes the proximal operator with respect to the data fidelity term, which is defined by
\begin{equation}
  \text{prox}_{\alpha_t \mathcal{D}}(\mathbf{\tilde u}) = \arg \min_{\mathbf{u}} \mathcal{D}(\mathbf{u}) + \frac{1}{2\alpha_t} \lVert \mathbf{u - \tilde u}\rVert^2_2.
  \label{eq:proximalOperator}
\end{equation}
Note that the proximal map allows to handle the non-smooth data fidelity terms such as the $\ell_1$ norm. 
Additionally, there is a strong link between proximal gradient methods and residual units which allows to incrementally reconstruct a solution (see \cref{fig:overview}).
We provide the computation of the prox-terms in the supplementary material.

\paragraph{Variational Network}
Our collaborative denoising algorithm consists of performing a fixed number of $T$ iterations of the proximal gradient method \cref{eq:PGM}.
In order to increase the flexibility we allow the model parameters to change in each iteration.
\begin{equation}
  \mathbf{u}_{t+1} = \text{prox}_{\alpha_t \mathcal{D}(\cdot, \theta_t)}(\mathbf{u}_t - \alpha_t \nabla \mathcal{R}(\mathbf{u}_t, \theta_t)), ~ 0 \leq t \leq T - 1
\end{equation}
Following \cite{chen2015learning,Kobler2017_GCPR} we parametrize the derivatives of the potential functions using Gaussian radial basis functions (RBF)
\begin{equation}
  \rho^{l,t}_k(s) = \beta_k^{l,t}\sum_{b=1}^B w_{k,b}^{l,t} \exp \left(- \frac{(s - \gamma_b)^2}{2 \sigma ^2} \right),
  \label{eq:rbf}
\end{equation}
where $\gamma_b$ are the means regularly sampled on the interval $[-3,3]$, $\sigma$ is the standard deviation of the Gaussian kernel and $\beta_k^{l,t}$ is a scaling factor.
The linear operators $K^{l,t}_k$ are implemented as multi-channel 2D convolutions with convolution kernels $\kappa_k^{l,t}$.
In summary, the parameters in each step are given by $\theta_t = \{ \kappa_k^{l,t}, \beta_k^{l,t}, w_{k,b}^{l,t}, \mu^t, \nu^t, \lambda^t, \alpha_t, \}$.

 \section{Computing Inputs}
\label{sec:computingInputs}
Our proposed refinement method can be applied to an arbitrary
stereo method, provided it comes along with a cost-volume, which is
the case for the majority of existing stereo methods.

\paragraph{Probability volume}
Assume we have given a cost-volume $v: \Omega \times \{0,\dots,D-1\}
\rightarrow \mathbb{R}$, where smaller costs mean a higher likelihood
of the respective disparity values.
In order to map the values onto probabilities $p: \Omega \times
\{0,\dots,D-1\}$, we make use of the ``softmax'' function, that is
\begin{equation}
  p(x, d) = \frac{\exp(\frac{ -v(x, d)}{\eta})}{\sum_{d'=0}^{D-1} \exp(\frac{-v(x, d')}{\eta})},
  \label{eq:softmax}
\end{equation}
where $\eta$ influences the smoothness of the probability distribution.

\paragraph{Initial disparity map}
From \cref{eq:softmax} we can compute the WTA solution by a pixel-wise arg max over the disparity dimension, i.e.,
\begin{equation}
  \bar d(x) \in \arg \max_d ~ p(x, d).
  \label{eq:wta}
\end{equation}
Moreover, we compute a sub-pixel accurate disparity map $\check d(x)$  by fitting a quadratic function to the probability volume. 
This is equivalent to perform one step of Newton's algorithm:
\begin{equation}
  \check d(x) = \bar d(x) - \frac{\delta^+(p(x, \cdot))(\bar d(x))}{\delta^{-} (\delta^{+}(p(x, \cdot)))(\bar d(x))},
  \label{eq:wta_refined}
\end{equation}
where $\delta^{\{+,-\}}$ denote standard forward and backward differences in the disparity dimension.
Furthermore, we compute the refined value of the probabilities, denoted as $\check p(x)$, via linear interpolation in the probability volume.

\paragraph{Initial Confidence Measure}
The computation of a confidence measure of the stereo results is important for many applications and a research topic on its own \cite{XuConfidences}.
Here we take advantage of the probabilistic nature of our matching costs $\check p(x)$.
Moreover, we make use of geometric constraints by using a left-right (LR) consistency check, where the left and right images are interchanged.
This allows us to identify occluded regions. 
We compute the probability of a pixel being not occluded as 
\begin{equation}
  p_{o}(x) = \frac{\max(\varepsilon - \text{dist}_{lr}(x), 0)}{\varepsilon} \in [0,1],
  \label{eq:conf_lr}
\end{equation}
where  
\begin{equation}
  \text{dist}_{lr}(x) = \lvert {\check d}_{l}(x) + {\check d}_{r}(x + \check d_{l}(x)) \rvert
\end{equation}
is the disparity difference between the left prediction $\check d_l$ and the right prediction $\check d_r$ and the parameter $\varepsilon$ acts as a threshold and is set to $\varepsilon = 3$ in all experiments.
The final confidence measure is given by
\begin{equation}
  c(x) = \check p(x) p_{o}(x) \in [0,1].
  \label{eq:conf_all}
\end{equation}
Thus, we define our total confidence as the product of the matching confidence and the LR confidence.
Most of the pixels not surviving the LR check are pixels in occluded regions.
To get a good initialization for these pixels as well, we inpaint the disparities of these pixels from the left side. 
The experiments show that this significantly increases the performance of the model (see \cref{tab:kittiAblation}).

 \section{Learning}
\label{sec:learning}
In this section we describe our learning procedure for the collaborative denoising model.
To remove scaling ambiguities we require the filter kernels $\kappa^{l,t}_k$ to be zero-mean and to have an $\ell_2$ norm $\leq 1$.
Moreover, we constraint the weights of the RBF kernels to have an $\ell_2$ norm $\leq 1$, too.
This is defined with the  following convex set: 
\begin{equation}
  \Theta = \{ \theta_t : \lVert \kappa^{l,t}_k \rVert \leq 1, ~ \sum \kappa^{l,t}_k = 0, ~ \lVert w_{k}^{l,t} \rVert \leq 1 \}
\end{equation}

For learning, we define a loss function that measures the error between the last iterate of the disparity map $u^d_T$ and the ground-truth disparity $d^*$.
Note that we do not have a loss function for the confidence and the color image.
Their aim is rather to support the disparity map to achieve the lowest loss.
We use a truncated Huber function of the form 
\begin{equation}
   \min_{\theta \in \Theta} \sum_{s=1}^S\sum_{i=1}^{MN} \min \left( \lvert u^d_{s,T}(x,\theta) - d_s^*(x) \rvert_\delta, ~\tau \right)
  \label{eq:loss}
\end{equation} 
where $\tau$ is a truncation value, $s$ denotes the index of the training sample and 
\begin{equation}
 \lvert r \rvert_\delta = 
  \begin{cases}
    \frac{r^2}{2\delta} & \text{if } \lvert r \rvert \leq \delta \\
    \lvert r \rvert - \frac{\delta}{2} & \text{else}
  \end{cases}
  \label{eq:huber}
\end{equation}
is the Huber function. 

\begin{figure*}[t!]
  \centering
  \begin{subfigure}{\textwidth}
    \centering
    \setlength{\tabcolsep}{0.5pt}
        \begin{tabular}{cccccccc}
       \textbf{Init} & \textbf{Step 1} & \textbf{Step 2} & \textbf{Step 3} & \textbf{Step 4} & \textbf{Step 5} & \textbf{Step 6} & \textbf{Step 7} \\
      \includegraphics[width=0.12\textwidth, clip, trim={0 150 250 0}]{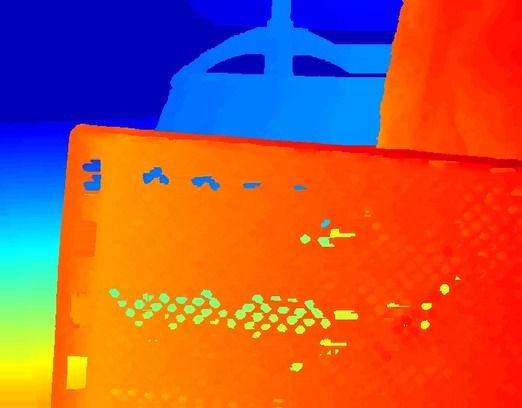} &
      \includegraphics[width=0.12\textwidth, clip, trim={0 150 250 0}]{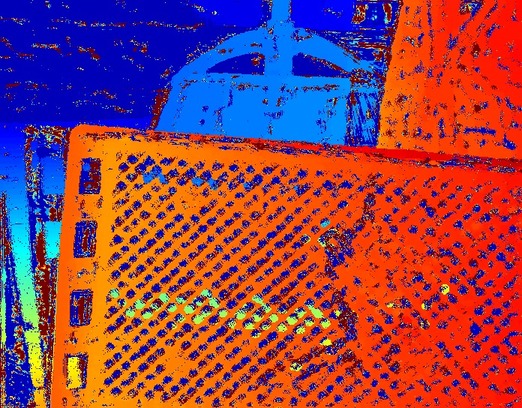} &
      \includegraphics[width=0.12\textwidth, clip, trim={0 150 250 0}]{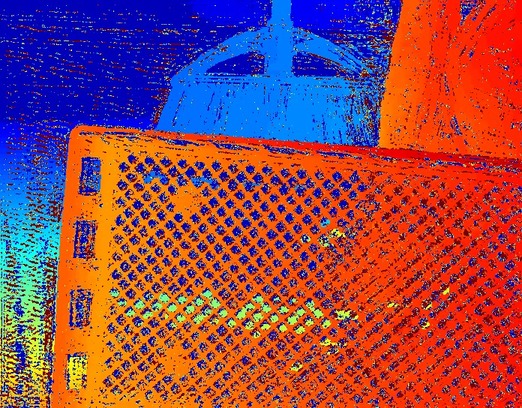} &
      \includegraphics[width=0.12\textwidth, clip, trim={0 150 250 0}]{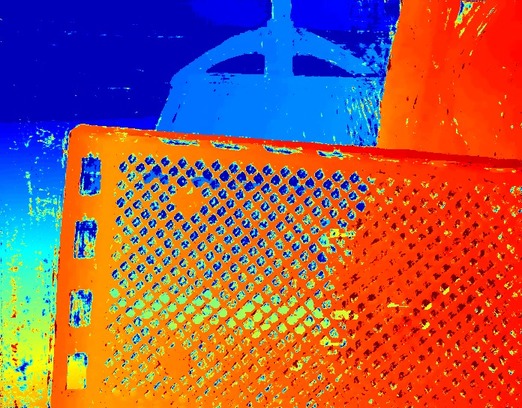} &
      \includegraphics[width=0.12\textwidth, clip, trim={0 150 250 0}]{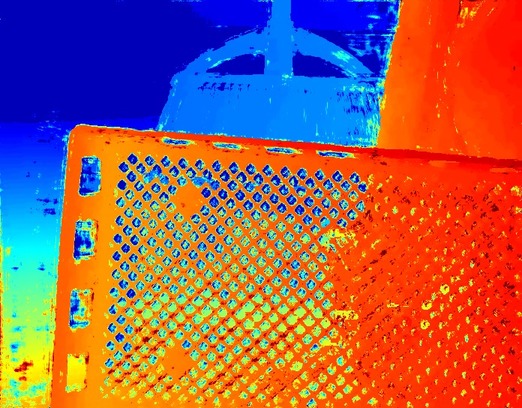} &
      \includegraphics[width=0.12\textwidth, clip, trim={0 150 250 0}]{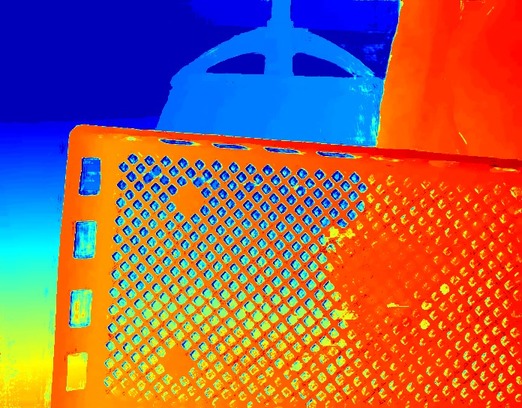} &
      \includegraphics[width=0.12\textwidth, clip, trim={0 150 250 0}] {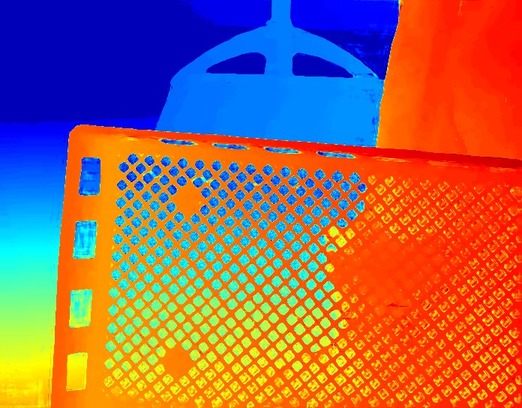} &
      \includegraphics[width=0.12\textwidth, clip, trim={0 150 250 0}]{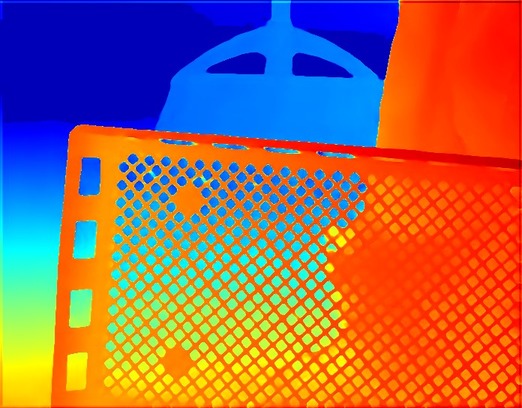} 
      \\
      \includegraphics[width=0.12\textwidth, clip, trim={0 150 250 0}]{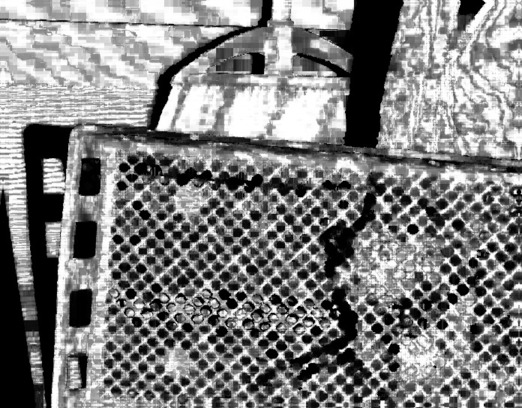} &
      \includegraphics[width=0.12\textwidth, clip, trim={0 150 250 0}]{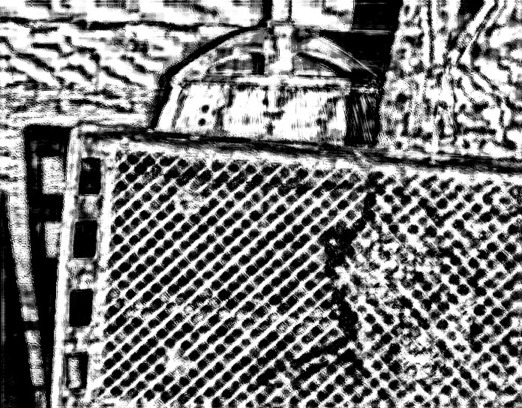} &
      \includegraphics[width=0.12\textwidth, clip, trim={0 150 250 0}]{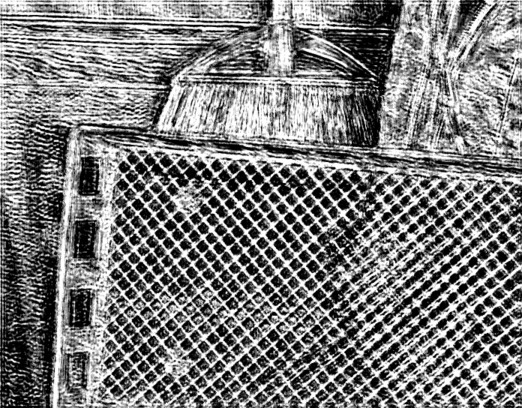} &
      \includegraphics[width=0.12\textwidth, clip, trim={0 150 250 0}]{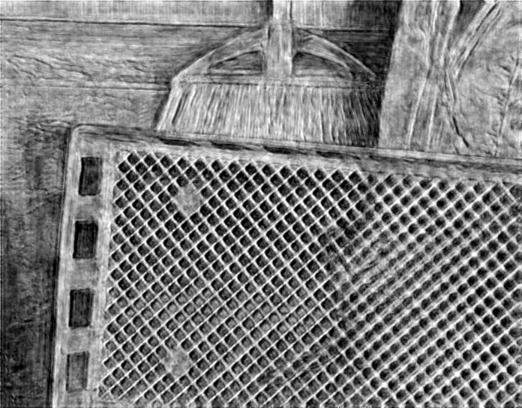} &
      \includegraphics[width=0.12\textwidth, clip, trim={0 150 250 0}]{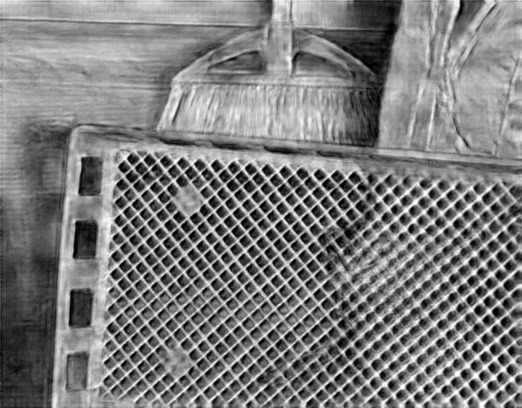} &
      \includegraphics[width=0.12\textwidth, clip, trim={0 150 250 0}]{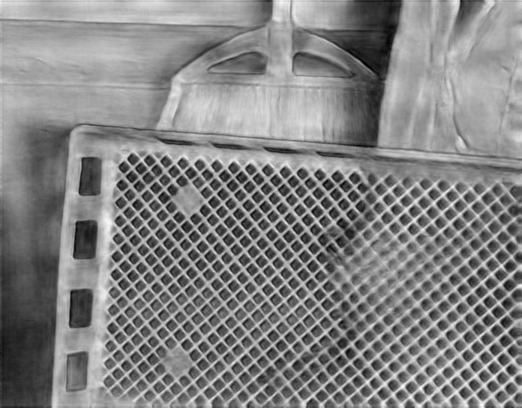} &
      \includegraphics[width=0.12\textwidth, clip, trim={0 150 250 0}]{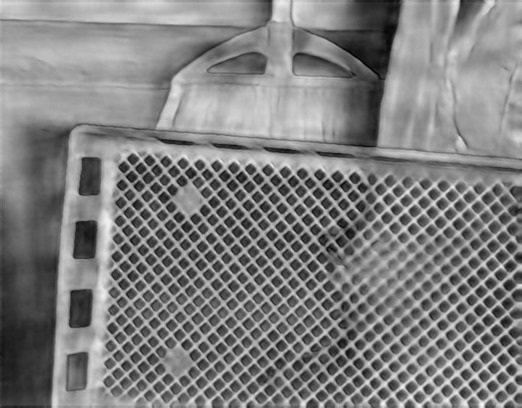} &
      \includegraphics[width=0.12\textwidth, clip, trim={0 150 250 0}]{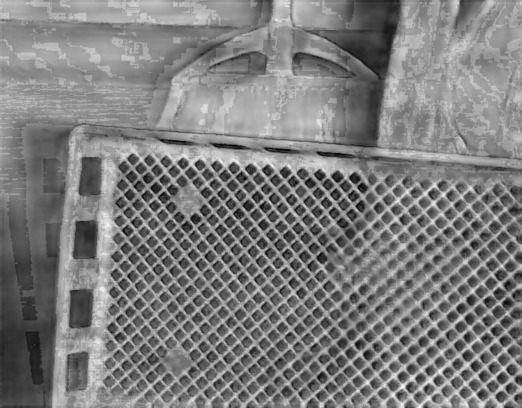} \\
      \includegraphics[width=0.12\textwidth, clip, trim={0 150 250 0}]{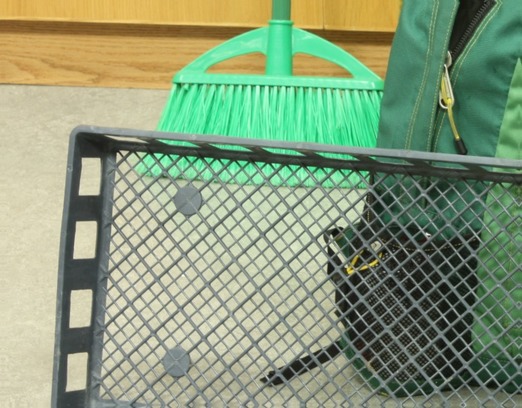} &
      \includegraphics[width=0.12\textwidth, clip, trim={0 150 250 0}]{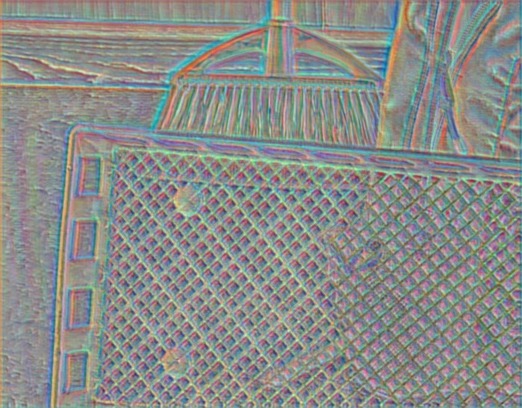} &
      \includegraphics[width=0.12\textwidth, clip, trim={0 150 250 0}]{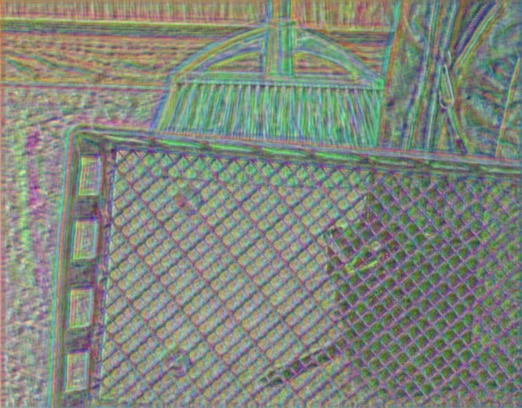} &
      \includegraphics[width=0.12\textwidth, clip, trim={0 150 250 0}]{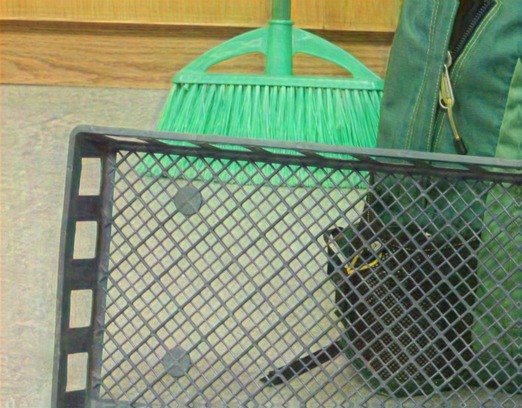} &
      \includegraphics[width=0.12\textwidth, clip, trim={0 150 250 0}]{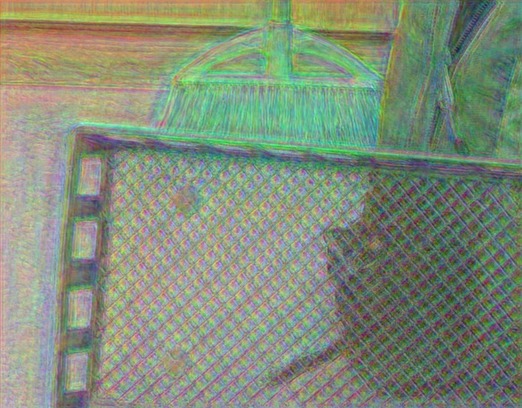} &
      \includegraphics[width=0.12\textwidth, clip, trim={0 150 250 0}]{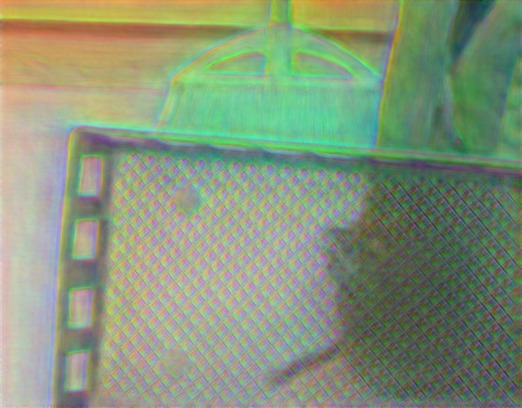} &
      \includegraphics[width=0.12\textwidth, clip, trim={0 150 250 0}]{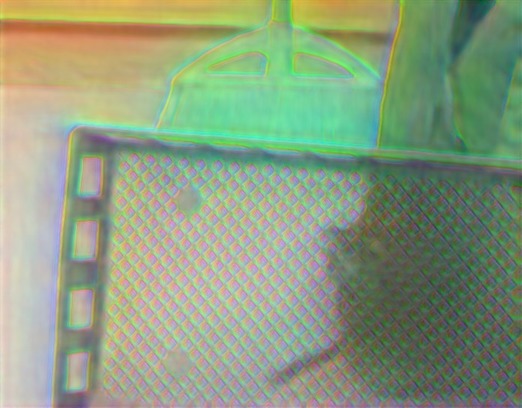} &
      \includegraphics[width=0.12\textwidth, clip, trim={0 150 250 0}]{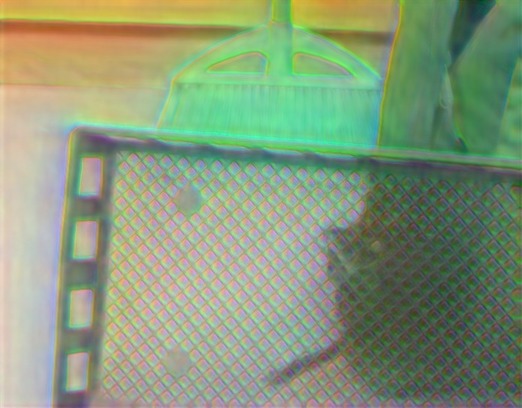} \\
    \end{tabular}
  \end{subfigure}
  \vspace{-0.5em}
  \caption{Visualization of steps in the VN. Top to bottom: disparity map, confidence map, image.  Left to right: Initialization, VN Steps 1 - 7. Note how the color image and the confidence map help to restore very fine details in the disparity map.
  }
  \label{fig:vnstepsvis}
\end{figure*} 
\paragraph{Implementation details}
We implemented our model in the PyTorch machine learning framework\footnote{https://pytorch.org}.
We train the refinement module for 3000 epochs with a learning rate of $10^{-3}$ with a modified projected Adam optimizer \cite{kingma2014adam}. We dynamically adjusted the stepsize computation to be constant within each parameter block in our constraint set $\Theta$ to ensure that we perform an orthogonal projection onto the constraint set. After 1500 epochs we reduce the truncation value $\tau$ from $\infty$ to $3$.
%
 \section{Experiments}
\label{sec:experiments}
We split the experiments into two parts.
In the first  part we evaluate architectural choices based on the WTA result of a matching network and compare with the FBS \cite{Barron2016}.
In the second part, we use the best architecture and train a variational network for refining the disparity maps computed by the CNN-CRF method \cite{Knobelreiter_2017_CVPR}. 
We use this method to participate in the public available stereo benchmarks Middlebury 2014 and Kitti 2015.
To ensure a fair comparison we choose methods with similar numbers of parameters and runtimes.
\cref{fig:vnstepsvis} shows how our method constructs the final result. 
The method recovers step-by-step fine details with the guidance of the confidences and the color image. 
\begin{figure}[t]
  \begin{subfigure}{0.48\columnwidth}
    \includegraphics[width=\textwidth]{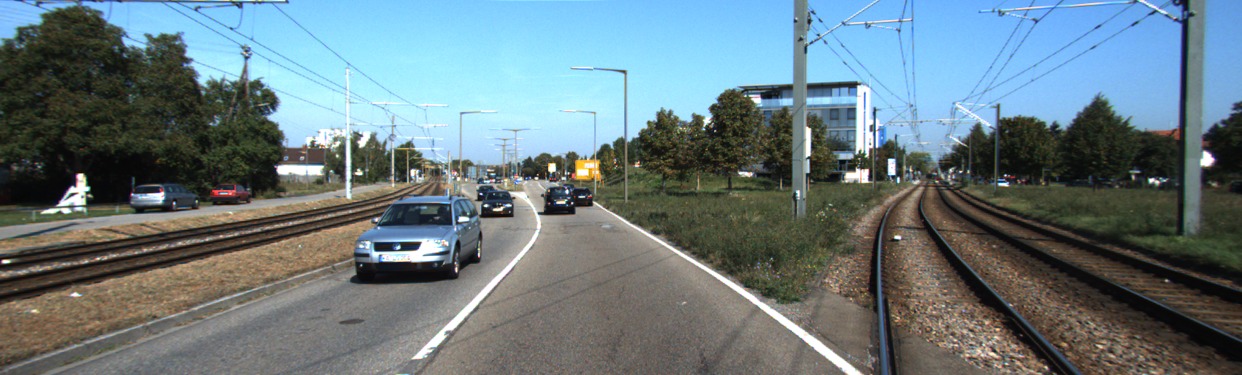}
    \includegraphics[width=\textwidth]{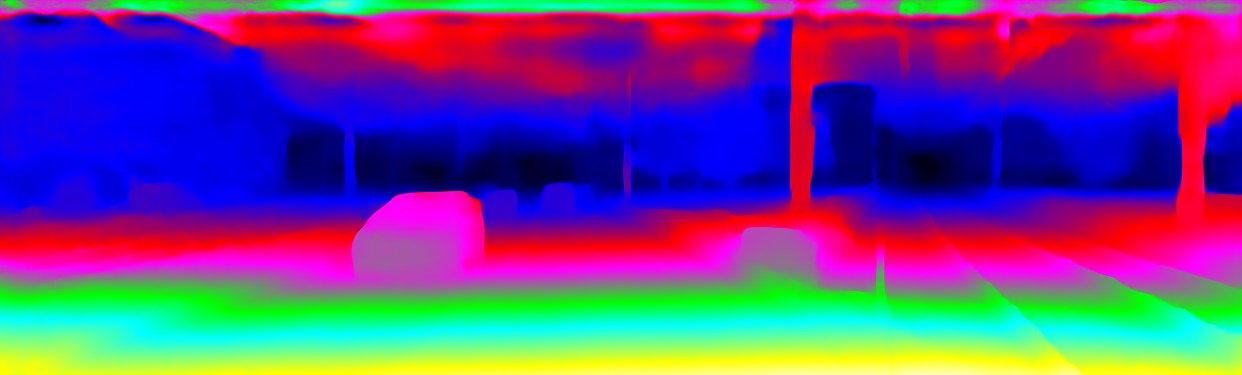}
    \includegraphics[width=\textwidth]{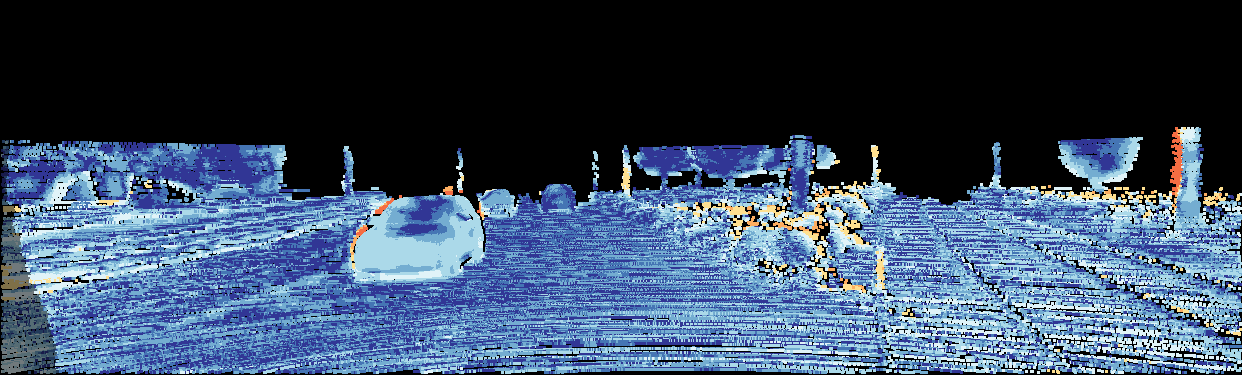}
  \end{subfigure}
  \begin{subfigure}{0.48\columnwidth}
    \includegraphics[width=\textwidth]{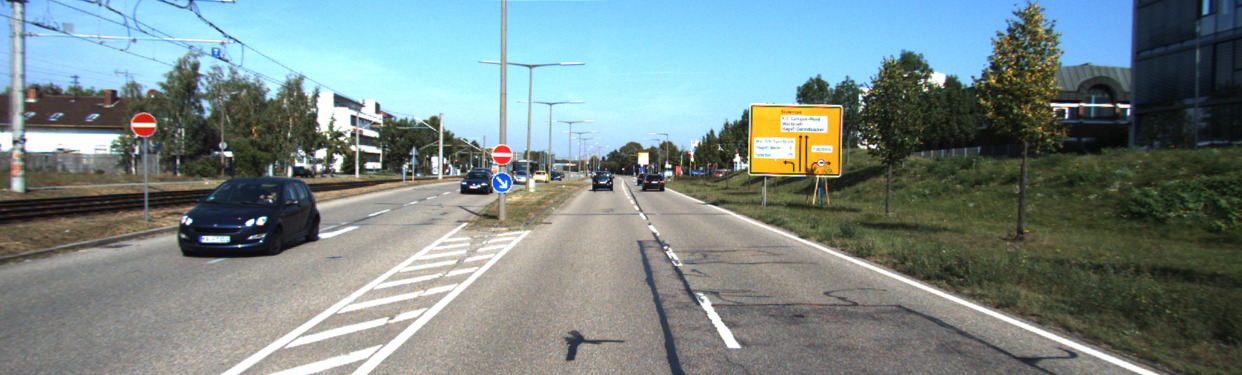}
    \includegraphics[width=\textwidth]{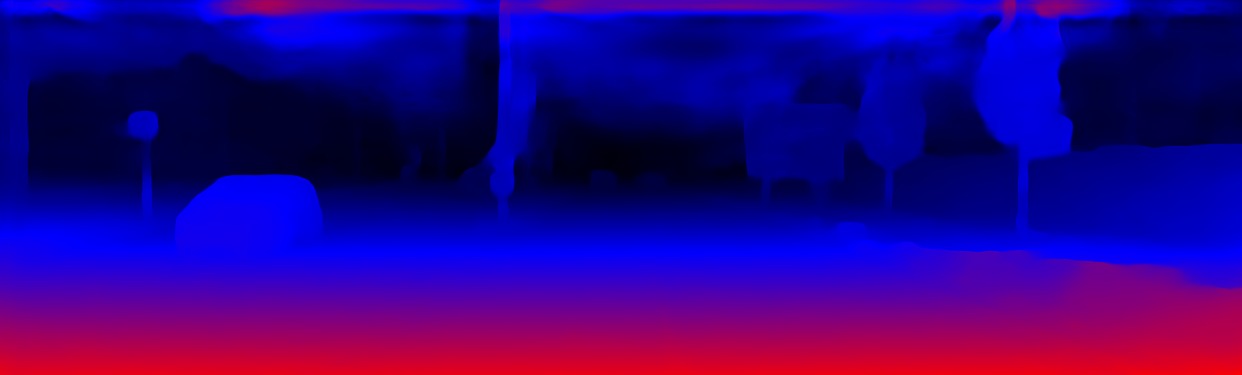}
    \includegraphics[width=\textwidth]{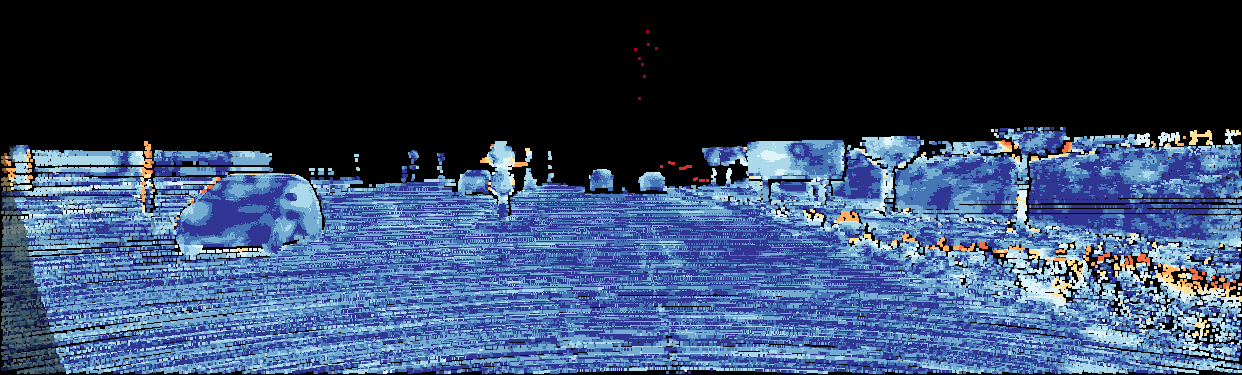}
  \end{subfigure}
  \caption{Qualitative results on the Kitti 2015 test set. Top-to-bottom: Reference image, disparity map which is color coded with blue = far away to yellow = near, error map, where blue = correct disparity, orange = incorrect disparity.}
  \label{fig:kittiTest}
\end{figure}

\paragraph{Kitti 2015}
\label{ssec:kitti}
The Kitti 2015 dataset \cite{Menze2015} is an outdoor dataset specifically designed for autonomous driving.
It contains 200 images with available ground-truth to train a model and 200 images with withheld ground-truth which is used for testing the models on previously unseen data.
The ground-truth is captured using a laser scanner and is therefore sparse in general.
The cars are densified by fitting CAD models into the laser point-cloud.
We report the {\em badX} error metric for occluded (occ) and non-occluded (noc) pixels with $X=3$. 
In the badX measure the predicted disparity $\hat d$ is treated incorrect, if the distance to the ground-truth disparity $d^*$ is larger than $X$. 

\paragraph{Middlebury 2014}
\label{ssec:middlebury}
The Middlebury 2014 stereo dataset \cite{Scharstein2014} is orthogonal to the Kitti 2015 dataset. 
It consists of 153 high resolution indoor images with highly precise dense ground-truth.
The challenges in the Middlebury dataset are large, almost untextured regions, huge occluded regions, reflections and difficult lighting conditions.
The generalization capability of the method is evaluated on a 15 images test-set with withheld ground-truth data.
We report all available metrics, i.e., bad$\{0.5,1,2,4\}$ errors, the average error (avg) and the root-mean-squared error (rms).

\subsection{Ablation Study}
\begin{table}[tb]
  \centering
  \begin{tabular}{cc|cc||cccccc||cccc}
    \toprule
    \multicolumn{2}{c|}{\textbf{Model}} & WTA & FBF \cite{Barron2016} & VN$^{7,5}_4$ & VN$^{7,5}_4$ & VN$^{7,5}_4$ & VN$^{7,5}_4$ & VN$^{7,5}_4$ & VN$^{7,5}_4$ & VN$^{5,7}_3$ & VN$^{8,7}_2$ & VN$^{14,3}_4$ & VN$^{11,3}_5$ \\
    \midrule
    \multicolumn{2}{c|}{\textbf{Conf}} & & \checkmark & & \checkmark & \checkmark & & \checkmark & \checkmark & \checkmark & \checkmark & \checkmark & \checkmark \\
    \multicolumn{2}{c|}{\textbf{Img}} & & \checkmark & &  \checkmark & &  \checkmark & \checkmark & \checkmark & \checkmark & \checkmark & \checkmark & \checkmark \\
    \multicolumn{2}{c|}{\textbf{OccIp}} & & \checkmark & \checkmark & &  \checkmark & \checkmark & \checkmark & \checkmark & \checkmark & \checkmark & \checkmark & \checkmark\\
    \multicolumn{2}{c|}{\textbf{Joint}} & & & & & & & &  \checkmark & \checkmark & \checkmark & \checkmark & \checkmark\\
    \midrule
    \multirow{2}{*}{\rotatebox[origin=c]{90}{\textbf{[bad3]}}} 
    & occ & 8.24 & 7.48 & 5.42 & 5.12 & 4.43 & 3.77 & 3.46 & \textbf{3.37} & 3.43 & 3.62 & 4.37 & 4.25 \\
    &noc & 6.78 & 6.08 & 4.68 & 3.98 & 3.90 & 3.07 & 2.72 & \textbf{2.55} & 2.58 & 2.97 & 3.71 & 3.49 \\
    \vspace{-8pt} \\
    \bottomrule
  \end{tabular}
  \caption{Ablation study on the Kitti 2015 dataset. 
           Conf = Confidences, Img = Image, OccIp = Occlusion inpainting, Joint = joint training.
            The super-script indicates the number of steps and the filter-size while the sub-script indicates the number of levels in the variational network. 
           VN$^{7,5}_4$ is therefore a variational network with $7$ steps and 4 levels.}
  \label{tab:kittiAblation}
\end{table}
To find the most appropriate hyper parameters for the proposed method, 
we generate our initial disparity map with a simple feature network.
The learned features are then compared using a fixed matching function for a pre-defined number of discrete disparities.

\paragraph{Feature Network}
\label{ssec:featureNet}

Our feature network is a modified version of the U-Net \cite{ronneberger2015u,long2015fully} which we use to extract features suitable for stereo matching. 
We kept the number of parameters low by only using 64 channels at every layer.
The output of our feature network is thus a 64-dimensional feature vector for every pixel. 
The exact architecture is shown in the supplementary material.

\paragraph{Feature Matching}
Next, we use the extracted features ${\psi^0}$ from the left image and ${\psi^1}$ from the right image to compute a matching score volume $\tilde p: \Omega \times \{0, \ldots, D-1\} \rightarrow \mathbb{R}$ with 
\begin{equation}
  \tilde p(x, d) = \langle \psi^0(x), ~ \psi^1(x - \tilde d) \rangle. 
  \label{eq:corr}
\end{equation}
We follow \cref{sec:computingInputs} to compute the inputs for the variational network. 

\paragraph{Ablation Study}
We systematically remove parts of our method in order to show how the final performance is influenced by the individual parts.
\cref{tab:kittiAblation} shows an overview of all experiments.
First we investigate the influence of our data-terms, the disparity data-term, the confidence data-term and the RGB image data-term.
The study shows that each of the data-terms positively influence the final performance.
Especially, adding the original input image significantly increases the performance. 
This can be \eg~seen in \cref{fig:vnstepsvis}, where the information of how the basket needs to be reconstructed, is derived from the input image.
In the second part of the study, we evaluate different variational network architectures.
To make the comparison as fair as possible, we chose the variants such that the total number of parameters is approximately the same for all architectures.
The experiments show, that a compromise between number of steps, pyramid levels and filter-size yields the best results.
The best performing model is the model $\text{VN}^{7,5}_4$, where the filter-size is set to $5 \times 5$ for $4$ pyramid levels and $7$ steps. The average runtime of this VN is as low as 0.09s on an NVidia 2080Ti graphics card.

Additionally, we compare with the FBS. 
We therefore use exactly the same inputs as we did in our method, i.e., the refined WTA solution $\check d$, our confidence measure $c$ and the RGB input image.
To ensure the best performance for the FBS, we performed a grid-search over its hyper-parameters on the Kitti dataset. 
As shown in \cref{tab:kittiAblation} the FBS clearly improves the performance upon the initial solution, but the FBS cannot compete with the proposed method.

\subsection{Benchmark Performance}
We use our method on top of the CNN-CRF~\cite{Knobelreiter_2017_CVPR} stereo method for the official test set evaluation (see \cref{tab:tabEval}).
We set the temperature parameter $\eta = 0.075$ in all experiments.
The average rank is computed with all published methods in the respective benchmarks with a runtime $\leq 20s$.
This yields in total 71 methods on the Kitti benchmark and 50 methods on the Middlebury benchmark.
\begin{table}[t]
  \centering
  \small
  \setlength\tabcolsep{4pt}
  \begin{tabular}{l|ccc|cccccccc}
    \toprule
    \multirow{2}{*}{\textbf{Method}} & \multicolumn{3}{c|}{\textbf{Kitti 2015}} & \multicolumn{7}{c}{\textbf{Middlebury 2014}} \\  
    & \textbf{noc} & \textbf{all}  & $\varnothing$\textbf{R} & \textbf{bad0.5} & \textbf{bad1} & \textbf{bad2} & \textbf{bad4} & \textbf{avg}  & \textbf{rms}  & \textbf{time} & $\varnothing$\textbf{R} \\
    \midrule
     PSMNet \cite{Chang_2018_CVPR}   &  
     \textbf{-} & \textbf{1.83} & - & 90.0 & 78.1 & 58.5 & 32.2 & 9.60 & 21.7 & 2.62 & 44\\
     PDS \cite{Tulyakov2018_NIPS} & - & - & - & 54.2 & 26.1 & 11.4 & 5.10 & 1.98 & 9.10 & 10 & 8\\
     \midrule
    MC-CNN \cite{Zbontar2016} & - & - & - & 42.1 & 20.5 & 11.7 & 7.94 & 3.87 & 16.5 & 1.26 & 9\\
     CNN-CRF \cite{Knobelreiter_2017_CVPR} & - & 4.04 & - & 56.1 & 25.1 & 10.8 & 6.12 & 2.30 & 9.89 & 3.53 & 10\\
    \midrule
    \cite{Knobelreiter_2017_CVPR} + VN (ours) & \textbf{\textit{1.90}} & \textit{2.04} & - & \textbf{\textit{41.8}} & \textbf{\textit{17.1}} & \textit{\textbf{7.05}} & \textbf{\textit{2.96}} & \textbf{\textit{1.21}} & \textbf{\textit{5.80}} & 4.06 & \textit{2}\\
    \bottomrule
    \midrule
     PSMNet \cite{Chang_2018_CVPR}   &  
     \textbf{2.14} & \textbf{2.32} & 17 & 81.1 & 63.9 & 42.1 & 23.5 & 6.68 & 19.4 & 2.62 & 33\\
     PDS \cite{Tulyakov2018_NIPS} & 2.36 & 2.58 & 19 & 58.9 & 21.1 & 14.2 & 6.98 & 3.27 & 15.7 & 10.3 & 9\\
     \midrule
     MC-CNN \cite{Zbontar2016}       & 
     3.33 & 3.89 & 32 &\textbf{41.3} & \textbf{18.0} & \textbf{9.47} & 6.7 & 4.37 & 22.4 & 1.26 & 6\\
    CNN-CRF \cite{Knobelreiter_2017_CVPR} & 4.84 & 5.50 & 36 & 60.9 & 31.9 & 12.5 & \textbf{6.61} & 3.02 & 14.4 & 3.53 & 8\\
    \midrule
    \cite{Knobelreiter_2017_CVPR} + VN (ours) & 
    \textit{4.45} & \textit{4.85} & \textit{33} & \textit{56.2} & \textit{30.0} & 14.2 & 7.71 & \textbf{\textit{2.49}} & \textbf{\textit{10.8}} & 4.06 & \textit{6}\\
    \bottomrule
  \end{tabular}
  \caption{Performance on public benchmarks. Top = Official training set, Bottom = Official test set. Bold font: Overall best. Italic font = improvement of base-line. $\varnothing$R denotes the average rank over all metrics on the benchmarks.}
  \label{tab:tabEval}
\end{table}

\begin{figure}[t]
  \centering
      \begin{tabular}{m{0.25cm}m{5.8cm}m{5.8cm}}
      \rotatebox[origin=t]{90}{\textbf{Initial}} 
      &
      \includegraphics[width=0.45\textwidth]{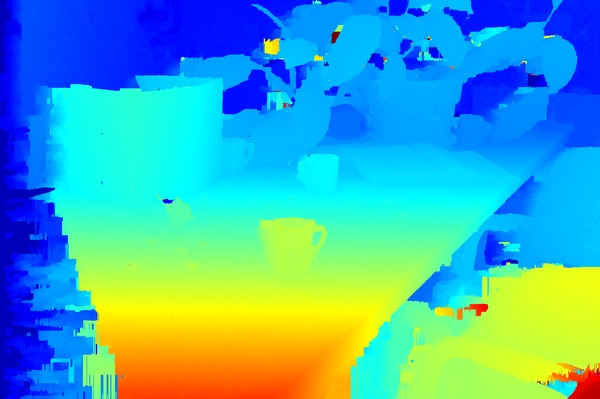} 
      &
      \includegraphics[width=0.45\textwidth]{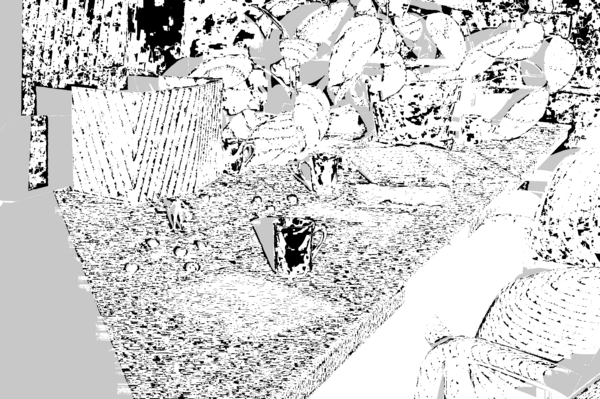} \\
      \rotatebox[origin=c]{90}{\textbf{Refined}}
      &
      \includegraphics[width=0.45\textwidth]{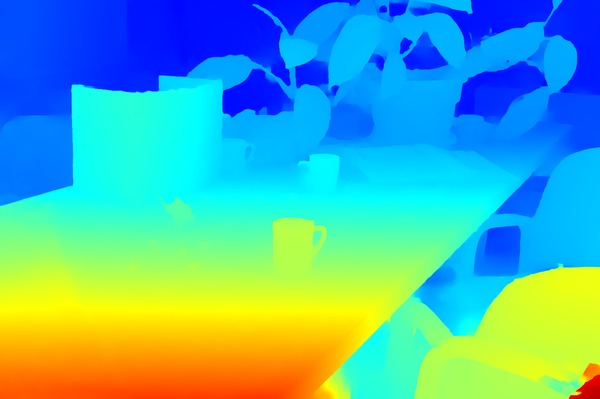}
      & 
     \includegraphics[width=0.45\textwidth]{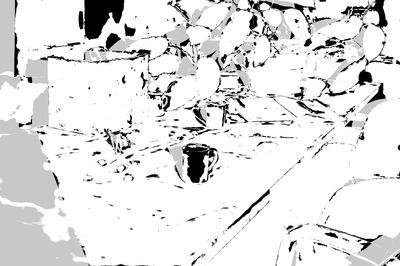}
    \end{tabular}
  \caption{Qualitative results on the Middlebury test set. Left: Color-coded disparity maps ranging from blue = far away to red = near. Right: Error maps, where white = correct and black = incorrect. The top row shows the initial disparity map (=input to the VN) and the bottom row shows our refined result.}
  \label{fig:mbTest}
\end{figure}
We used the model VN$^{7,5}_4$ on the Kitti dataset. As shown in \cref{tab:tabEval} we reduce the bad3 error in both, occluded and in non-occluded regions. \cref{fig:kittiTest} shows qualitative results with the corresponding error maps on the Kitti test set.

On the Middlebury benchmark we use the model VN$^{7,11}_4$ for all evaluations. 
We compare the errors on the training set with the errors on the test set (\cref{tab:tabEval}) and observe that:
(i) Our method achieves the best scores in all metrics among the compared methods on the training set. 
(ii) This positive trend is transferred to the test set for the average error and the RMS error. (iii) The bad$\{0.5,1\}$ errors on the test set are reduced and (iv) the bad$\{2,4\}$ errors slightly increase on the test set compared to \cite{Knobelreiter_2017_CVPR}.
One reason for this is the limited amount of training data for these very high-resolution images.
\cref{fig:mbTest} shows a qualitative example of the Middlebury test set. 
Note that the tabletop is nice and smooth while the sharp edges of the objects are very well preserved.

 \section{Conclusion \& Future Work}
\label{sec:conclusion}

We have proposed a learnable variational network for efficient refinement of disparity maps. 
The learned collaborative and hierarchical refinement method allows the use of information from the joint color, confidence and disparity space from multiple spatial resolutions. 
In an ablation study, we evaluated a broad range of architectural choices and demonstrated the impact of our design decisions. 
Our method can be applied on top of any other stereo method, with full use of additional information of a full cost volume.
We demonstrated this by adding the variational refinement network on top of the CNN-CRF method and have shown improved results. 
Furthermore, we have proven the effectiveness of our method by participating in the publicly available stereo benchmarks of Middlebury and Kitti. 
In future work, we would like to include a matching score during the refinement process and perform data augmentation to increase the training set for learning.

 
 \vspace{.75em}
 \noindent
 \textbf{Acknowledgements.} 
 This work was partly supported from the ERC starting grant HOMOVIS (No.640156). 

 	\bibliographystyle{splncs04}
 	\bibliography{newbib.bib}

\begin{thebibliography}{10}
\providecommand{\url}[1]{\texttt{#1}}
\providecommand{\urlprefix}{URL }
\providecommand{\doi}[1]{https://doi.org/#1}

\bibitem{Barron2016}
Barron, J.T., Poole, B.: The fast bilateral solver. In: European Conference on
  Computer Vision (ECCV). pp. 617--632 (2016)

\bibitem{beck2009fast}
Beck, A., Teboulle, M.: A fast iterative shrinkage-thresholding algorithm for
  linear inverse problems. {SIAM} J. Imaging Sciences pp. 183--202 (2009)

\bibitem{Brox2004_ECCV}
Brox, T., Bruhn, A., Papenberg, N., Weickert, J.: High accuracy optical flow
  estimation based on a theory for warping. In: European Conference on Computer
  Vision (ECCV). pp. 25--36 (2004)

\bibitem{chambolle2011first}
Chambolle, A., Pock, T.: A first-order primal-dual algorithm for convex
  problems with applications to imaging. Journal of Mathematical Imaging and
  Vision pp. 120--145 (2011)

\bibitem{Chang_2018_CVPR}
Chang, J.R., Chen, Y.S.: Pyramid stereo matching network. In: IEEE Conference
  on Computer Vision and Pattern Recognition (CVPR). pp. 5410--5418 (2018)

\bibitem{chen2015learning}
Chen, Y., Yu, W., Pock, T.: On learning optimized reaction diffusion processes
  for effective image restoration. In: IEEE Conference on Computer Vision and
  Pattern Recognition (CVPR). pp. 5261--5269 (2015)

\bibitem{gidaris2017detect}
Gidaris, S., Komodakis, N.: Detect, replace, refine: Deep structured prediction
  for pixel wise labeling. In: IEEE Conference on Computer Vision and Pattern
  Recognition (CVPR). pp. 5248--5257 (2017)

\bibitem{He_2016_CVPR}
He, K., Zhang, X., Ren, S., Sun, J.: Deep residual learning for image
  recognition. In: IEEE Conference on Computer Vision and Pattern Recognition
  (CVPR). pp. 770--778 (June 2016)

\bibitem{XuConfidences}
Hu, X., Mordohai, P.: A quantitative evaluation of confidence measures for
  stereo vision. IEEE Transactions on Pattern Analysis and Machine Intelligence
  pp. 2121--2133 (2012)

\bibitem{kingma2014adam}
Kingma, D.P., Ba, J.: Adam: A method for stochastic optimization. arXiv
  preprint arXiv:1412.6980  (2014)

\bibitem{Knobelreiter_2017_CVPR}
Kn\"obelreiter, P., Reinbacher, C., Shekhovtsov, A., Pock, T.: End-to-end
  training of hybrid cnn-crf models for stereo. In: IEEE Conference on Computer
  Vision and Pattern Recognition (CVPR). pp. 2339--2348 (2017)

\bibitem{Kobler2017_GCPR}
Kobler, E., Klatzer, T., Hammernik, K., Pock, T.: Variational networks:
  Connecting variational methods and deep learning. In: German Conference on
  Pattern Recognition (GCPR). pp. 281--293 (2017)

\bibitem{kuschk2013fast}
Kuschk, G., Cremers, D.: Fast and accurate large-scale stereo reconstruction
  using variational methods. In: IEEE International Conference on Computer
  Vision Workshop. pp. 700--707 (2013)

\bibitem{Liang_2018_CVPR}
Liang, Z., Feng, Y., Guo, Y., Liu, H., Chen, W., Qiao, L., Zhou, L., Zhang, J.:
  Learning for disparity estimation through feature constancy. In: IEEE
  Conference on Computer Vision and Pattern Recognition (CVPR). pp. 2811--2820
  (2018)

\bibitem{long2015fully}
Long, J., Shelhamer, E., Darrell, T.: Fully convolutional networks for semantic
  segmentation. In: IEEE Conference on Computer Vision and Pattern Recognition
  (CVPR). pp. 3431--3440 (2015)

\bibitem{Maurer17}
Maurer, D., Stoll, M., Bruhn, A.: Order-adaptive and illumination-aware
  variational optical flow refinement. In: British Machine Vision Conference
  (2017)

\bibitem{Menze2015}
Menze, M., Geiger, A.: Object scene flow for autonomous vehicles. In: IEEE
  Conference on Computer Vision and Pattern Recognition (CVPR). pp. 3061--3070
  (2015)

\bibitem{Pang_2017_ICCV_Workshops}
Pang, J., Sun, W., Ren, J.S., Yang, C., Yan, Q.: Cascade residual learning: A
  two-stage convolutional neural network for stereo matching. In: IEEE
  International Conference on Computer Vision Workshop. pp. 887--895 (2017)

\bibitem{Neal_FTO_2014}
Parikh, N., Boyd, S., et~al.: Proximal algorithms. Foundations and
  Trends{\textregistered} in Optimization pp. 127--239 (2014)

\bibitem{Ranftl2012_IVS}
Ranftl, R., Gehrig, S., Pock, T., Bischof, H.: Pushing the limits of stereo
  using variational stereo estimation. In: IEEE Intelligent Vehicles Symposium.
  pp. 401--407 (2012)

\bibitem{ranftl2014non}
Ranftl, R., Bredies, K., Pock, T.: Non-local total generalized variation for
  optical flow estimation. In: European Conference on Computer Vision (ECCV).
  pp. 439--454 (2014)

\bibitem{RevaudWHS15}
Revaud, J., Weinzaepfel, P., Harchaoui, Z., Schmid, C.: Epicflow:
  Edge-preserving interpolation of correspondences for optical flow. In: IEEE
  Conference on Computer Vision and Pattern Recognition (CVPR). pp. 1164--1172
  (2015)

\bibitem{riegler2016atgv}
Riegler, G., R{\"u}ther, M., Bischof, H.: Atgv-net: Accurate depth
  super-resolution. In: European Conference on Computer Vision (ECCV). pp.
  268--284 (2016)

\bibitem{ronneberger2015u}
Ronneberger, O., Fischer, P., Brox, T.: U-net: Convolutional networks for
  biomedical image segmentation. In: International Conference on Medical Image
  Computing and Computer-Assisted Intervention (MICCAI). pp. 234--241 (2015)

\bibitem{Roth2009_FoE}
Roth, S., Black, M.J.: Fields of experts. International Journal of Computer
  Vision  (2009)

\bibitem{Scharstein2014}
Scharstein, D., Hirschm{\"u}ller, H., Kitajima, Y., Krathwohl, G., Nesic, N.,
  Wang, X., Westling, P.: High-resolution stereo datasets with
  subpixel-accurate ground truth. In: German Conference on Pattern Recognition
  (GCPR). pp. 31--42 (2014)

\bibitem{Scharstein2002}
Scharstein, D., Szeliski, R.: A taxonomy and evaluation of dense two-frame
  stereo correspondence algorithms. International Journal of Computer Vision
  (2002)

\bibitem{shekhovtsov2016solving}
Shekhovtsov, A., Reinbacher, C., Graber, G., Pock, T.: Solving dense image
  matching in real-time using discrete-continuous optimization. Computer Vision
  Winter Workshop  (2016)

\bibitem{Tulyakov2018_NIPS}
Tulyakov, S., Ivanov, A., Fleuret, F.: Practical deep stereo (pds): Toward
  applications-friendly deep stereo matching. In: Proceedings of Advances in
  Neural Information Processing Systems. pp. 5871--5881 (2018)

\bibitem{vogel2018learning}
Vogel, C., Kn{\"o}belreiter, P., Pock, T.: Learning energy based inpainting for
  optical flow. In: Asian Conference on Computer Vision (ACCV) (2018)

\bibitem{vogel2017primal}
Vogel, C., Pock, T.: A primal dual network for low-level vision problems. In:
  German Conference on Pattern Recognition (GCPR) (2017)

\bibitem{zach2007duality}
Zach, C., Pock, T., Bischof, H.: A duality based approach for realtime tv-l1
  optical flow. In: German Conference on Pattern Recognition (GCPR) (2007)

\bibitem{Zbontar2016}
{\v{Z}}bontar, J., LeCun, Y.: Stereo matching by training a convolutional
  neural network to compare image patches. Journal of Machine Learning Research
   (2016)

\end{thebibliography}

 	\newpage
 
\setcounter{figure}{0}
\setcounter{table}{0}
\setcounter{section}{0}

 {\centering
 \Large Learned Collaborative Stereo Refinement \\ Supplementary Material \\[1.5em] 
 \normalsize}

\section{Proximal Operators for the data terms}
We provide the solution of the proximal operators used in the data terms.
The proximal operator is defined as the optimization problem 
\begin{equation}
  \text{prox}_{\alpha f}(\tilde u) = \arg \min_{u} f(u) + \frac{1}{2 \alpha} \lVert u - \tilde u \rVert^2.
  \label{eq:prox_suppl}
\end{equation}
In the main paper we used the proximal operator with $\ell_1$ and $\ell_2$ functions. The $\ell_2$ prox is used for the color image and the $\ell_1$ prox for the confidences and the disparities.
We will present the closed form solutions for these two cases in the next paragraphs.

\paragraph{Proximal operator for $\ell_2$ functions}
First, we present the result of the proximal operator for the following $\ell_2$ function
\begin{equation}
  f(u) =  \frac{\lambda}{2} \lVert u - u_0 \rVert^2.
  \label{eq:ell2}
\end{equation}
Inserting \cref{eq:ell2} into \cref{eq:prox_suppl} and setting the derivative w.r.t. $u$ to zero, we can compute the optimal solution $u^*$ with 
\begin{equation}
  u^* = \frac{\tilde u + \tau \alpha u_0}{1 + \lambda \alpha},
\end{equation}
where for the color image data term, $u_0 = I_0$ and $\tilde u = u^{rgb}$.

\paragraph{Proximal operator for $\ell_1$ functions}
Similarly, we compute the proximal operator of the following weighted $\ell_1$ function
\begin{equation}
  f(u) =  \gamma \lVert u - u_0 \rVert_{w,1} = \gamma \sum_{x \in \Omega} w(x) \lvert u(x) - u_0(x) \rvert.
  \label{eq:ell1}
\end{equation}
The absolute function is not differentiable at 0 and 
therefore the optimality condition requires the sub-differential to contain 0.
The closed form solution of the proximal operator \cref{eq:prox_suppl} with $f$ being the $\ell_1$ function as defined in \cref{eq:ell1} is given by
\begin{equation}
  u^* = u_0 + \max (0, \lvert \tilde u - u_0 \rvert - \tau \gamma w) \text{sign}(\tilde u - u_0).
  \label{eq:prox_l1}
\end{equation}
Thus, for the confidence data term $w = 1$ and $u_0 = c$ and for the disparity data term $w = c$ and $u_0 = \check d$.

\section{Derivative of the quadratic fitting}
\begin{figure}[h]
  \centering
  \includegraphics[width=0.7\textwidth]{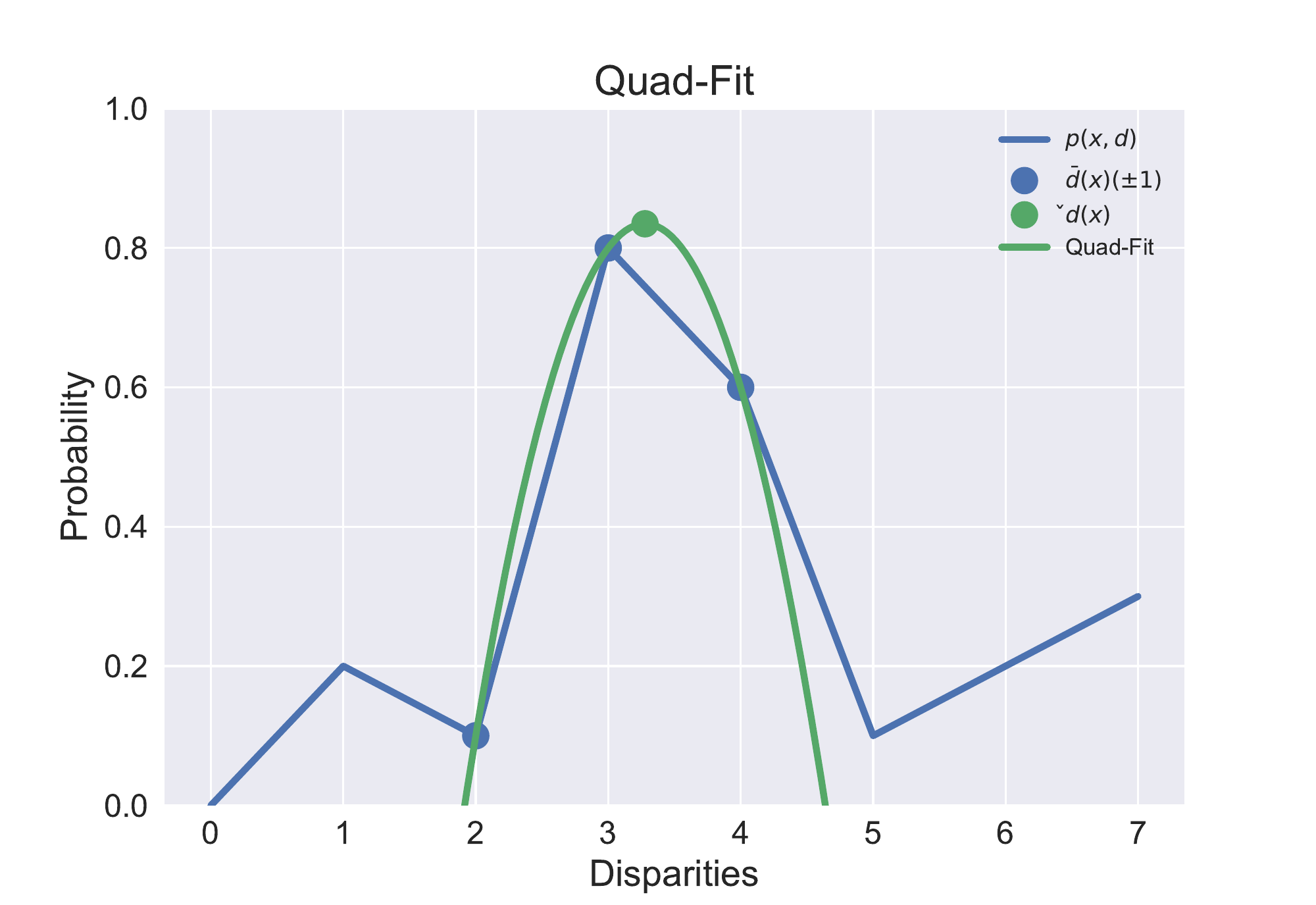}
  \caption{Visualization of the quadratic fitting. We select the points next to the maximum value and fit a quadratic function. Computing the extremum of the quadratic functions yields the refined disparity and the refined probability.}
  \label{fig:quadfit}
\end{figure}

In the joint training of our feature network and the regularization network we need to backpropagate the gradient through the refined disparities.
Therefore, we must compute the gradient of our sub-pixel accurate disparity map w.r.t. the probability volume.
The gradient is non-zero only for the supporting points of the quadratic function (shown in blue in \cref{fig:quadfit}).
Precisely, it is given by
\begin{equation}
  \frac{\partial \check d(x)}{\partial p(x, d)} = 
  \begin{cases}
    \frac{\delta^c(p(x,\cdot))(\bar d(x))}{(\delta^-(\delta^+(p(x,\cdot)))(\bar d(x)))^2} & \text{if } d = \bar d(x) \\[10pt]
    \frac{\delta^+(p(x,\cdot))(\bar d(x))}{(\delta^-(\delta^+(p(x,\cdot)))(\bar d(x)))^2} & \text{if } d = \bar d(x) - 1 \\[10pt]
    \frac{\delta^-(p(x,\cdot))(\bar d(x))}{(\delta^-(\delta^+(p(x,\cdot)))(\bar d(x)))^2} & \text{if } d = \bar d(x) + 1 \\[10pt]
    0 & \text{else},
  \end{cases}
\end{equation}
where $\delta^{\{+,-,c\}}$ are standard forward-, backward- and central-differences in the disparity dimension.
Note, that we overcome the problem of the non-differentiable $\arg \min$ function with the fitting of the quadratic function.
\cref{fig:quadfit} shows a visualization of the quadratic fitting procedure.

\section{Architecture of the feature network}
The detailed architecture of our feature network is shown in \cref{tab:featurenet}.
It is a variant of a U-Net with two spatial resolutions and 64 channels in every layer.
In the contraction part the resolution is reduced every 2 layers.
In the expansion part features are upsampled with transposed convolutions and concatenated with the extracted features of the same resolution to keep fine details.

\begin{table}[t]
  \centering
  \renewcommand{\arraystretch}{1.3}
  \scriptsize
  \begin{tabular}{ccccc}
    \toprule
    \textbf{Layer} & \textbf{KS} & \textbf{Resolution} & \textbf{Channels} & \textbf{Input} \\
    \midrule
    conv00 & 3 & $W \times H ~/ ~ W \times H$ & $3 ~/ ~64$ & Image \\
    conv01 & 3 & $W \times H ~/ ~ W \times H$ & $64~/~64$ & conv00 \\
    pool0 & 2 & $W \times H ~/ ~ \frac{W}{2} \times \frac{H}{2}$ & $64~/~64$ & conv01\\
    \midrule
    conv10 & 3 & $\frac{W}{2} \times \frac{H}{2} ~/~ \frac{W}{2} \times \frac{H}{2}$ & $64 ~/~64$ & pool0 \\
    conv11 & 3 & $\frac{W}{2} \times \frac{H}{2} ~/~ \frac{W}{2} \times \frac{H}{2}$ & $64 ~/~64$ & conv10 \\
    pool1 & 2 & $\frac{W}{2} \times \frac{H}{2} ~/~ \frac{W}{4} \times \frac{H}{4}$ & $64~/~64$ & conv10\\
    \midrule
    conv20 & 3 & $\frac{W}{4} \times \frac{H}{4} ~/~ \frac{W}{4} \times \frac{H}{4}$ & $64 ~/~64$ & pool1 \\
    conv21 & 3 & $\frac{W}{4} \times \frac{H}{4} ~/~ \frac{W}{4} \times \frac{H}{4}$ & $64 ~/~64$ & conv20 \\
    \midrule
    deconv1 & 3 & $\frac{W}{4} \times \frac{H}{4} ~/~ \frac{W}{2} \times \frac{H}{2}$ & $64 ~/~64$ & conv21 \\
    conv12 & 3 & $\frac{W}{2} \times \frac{H}{2} ~/~ \frac{W}{2} \times \frac{H}{2}$ & $128~/~64$ & \{deconv1, conv11\} \\
    conv13 & 3 & $\frac{W}{2} \times \frac{H}{2} ~/~ \frac{W}{2} \times \frac{H}{2}$ & $64 ~/~64$ & conv12 \\
    \midrule
    deconv0 & 3 & $\frac{W}{2} \times \frac{H}{2} ~/~ W \times H$ & $64 ~/~64$ & conv12 \\
    conv02 & 3 & $W \times H ~/~ W\times H$ & $128 ~/~64$ & \{deconv0, conv01\} \\
    conv03 & 3 & $W \times H ~/~ W \times H$ & $64 ~/~64$ & conv02 \\
    \bottomrule
  \end{tabular}
  \caption{Detailed architecture of our multi-level feature network. {\em KS} denotes the kernel size, {\em Resolution} contains the spatial resolution of the input and output, respectively and {\em Channels} contain the number of input and output feature channels, respectively. We use curly brackets to indicate a concatenation of feature maps.}
  \label{tab:featurenet}
\end{table}

\section{Learned filters and activation functions}
\begin{figure}[t]
  \centering
  \begin{subfigure}{0.45\textwidth}
    \includegraphics[width=\textwidth, clip, trim={0 380 190 0}]{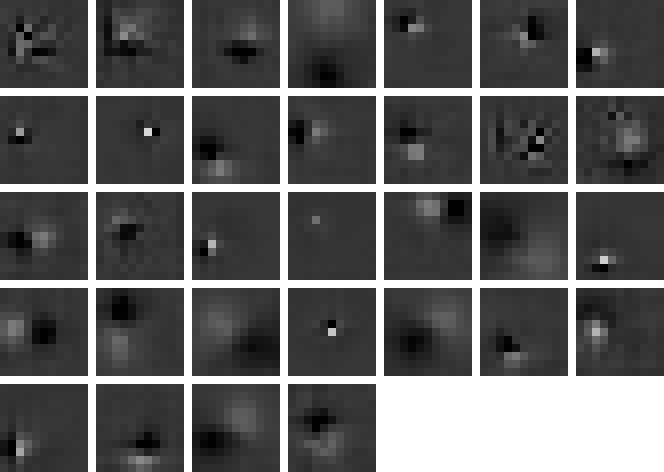}
    \includegraphics[width=\textwidth, clip, trim={0 90 190 290}]{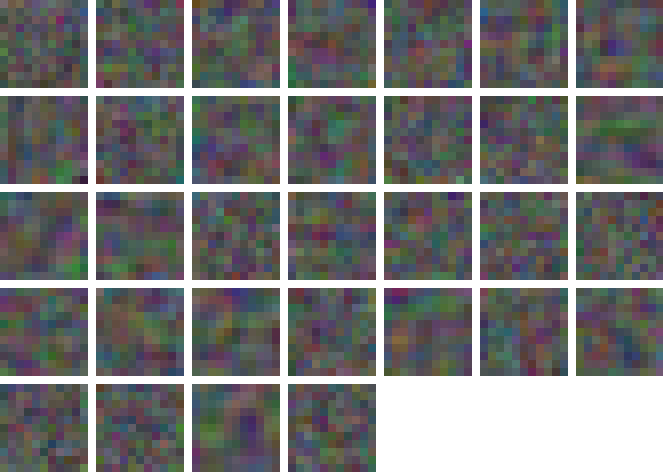}
    \includegraphics[width=\textwidth, clip, trim={96 90 95 290}]{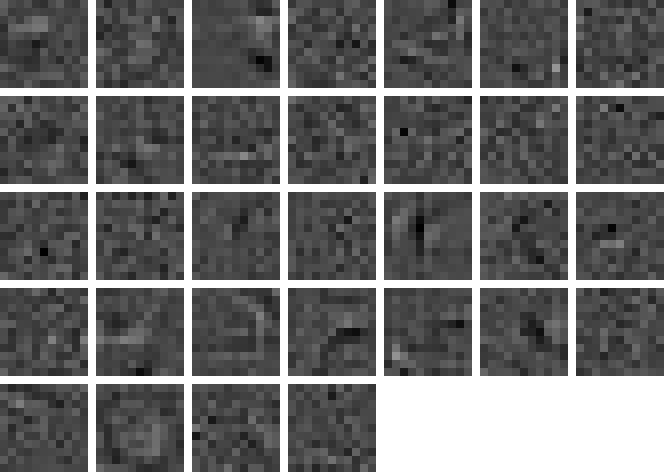}
    \caption{Filters}   
    \label{sfig:filters}
  \end{subfigure}
  ~
  \begin{subfigure}{0.5\textwidth}
    \includegraphics[width=0.24\textwidth]{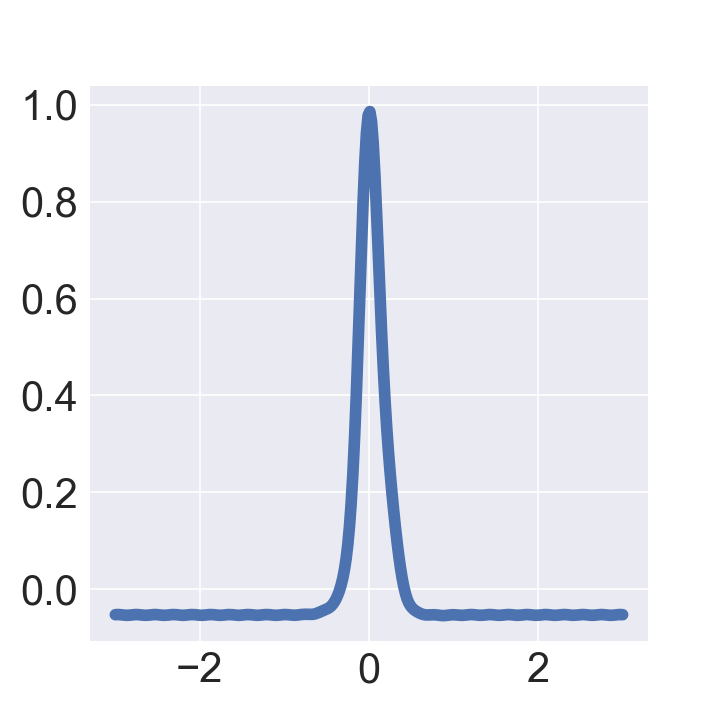}
    \includegraphics[width=0.24\textwidth]{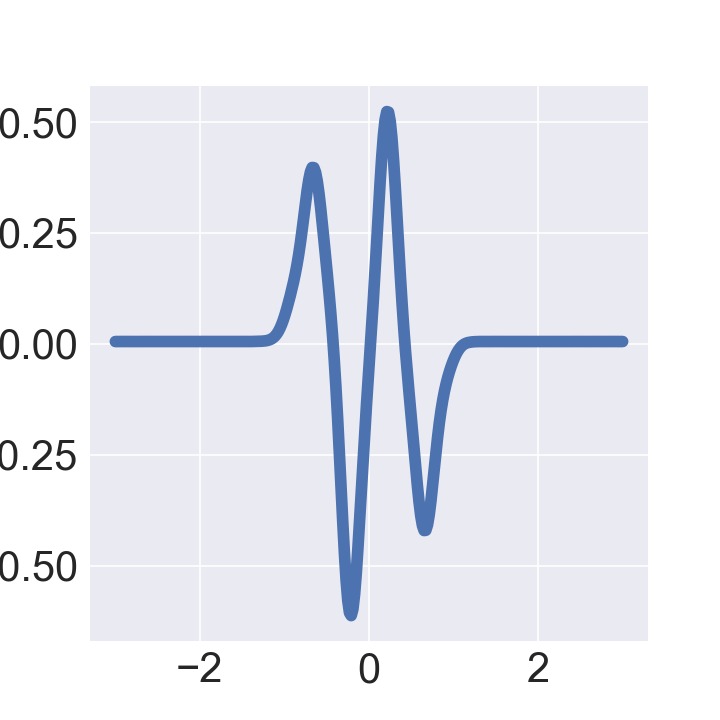}
    \includegraphics[width=0.24\textwidth]{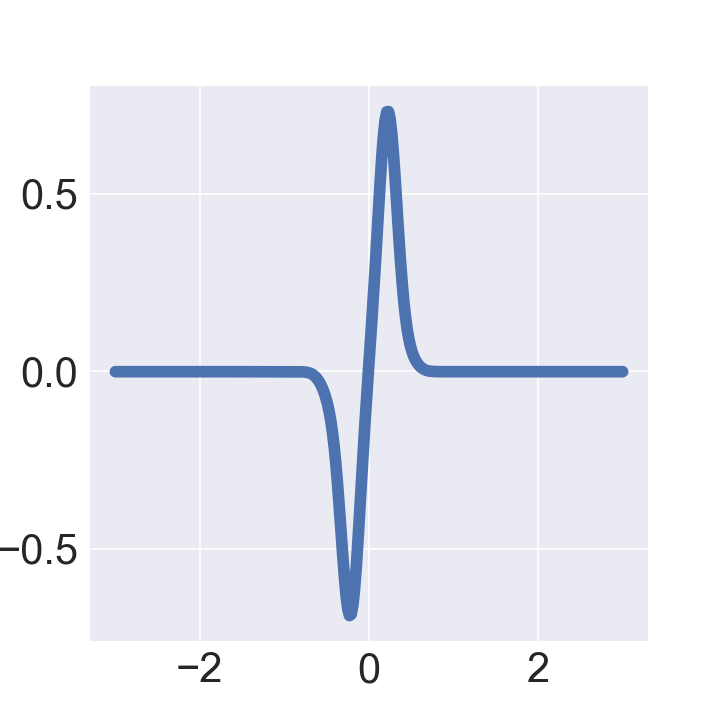}
    \includegraphics[width=0.24\textwidth]{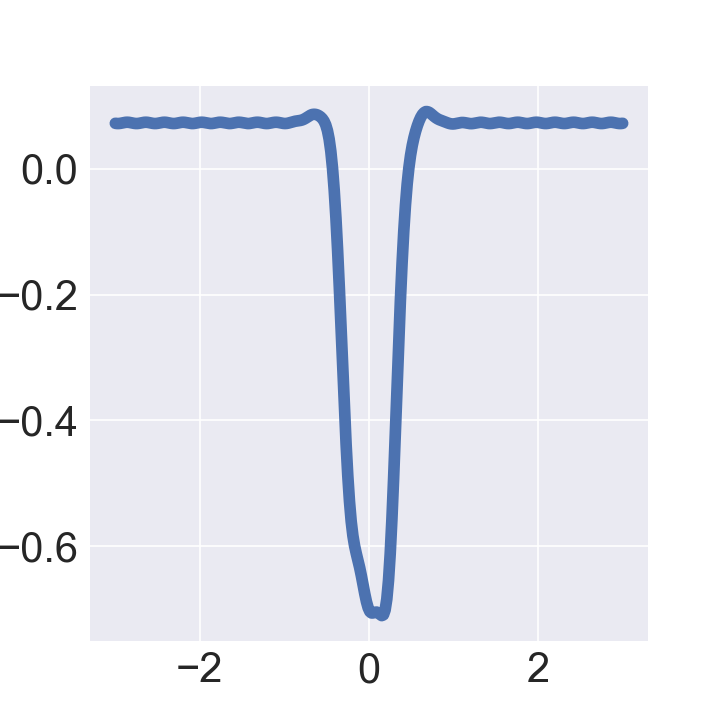}
    \includegraphics[width=0.24\textwidth]{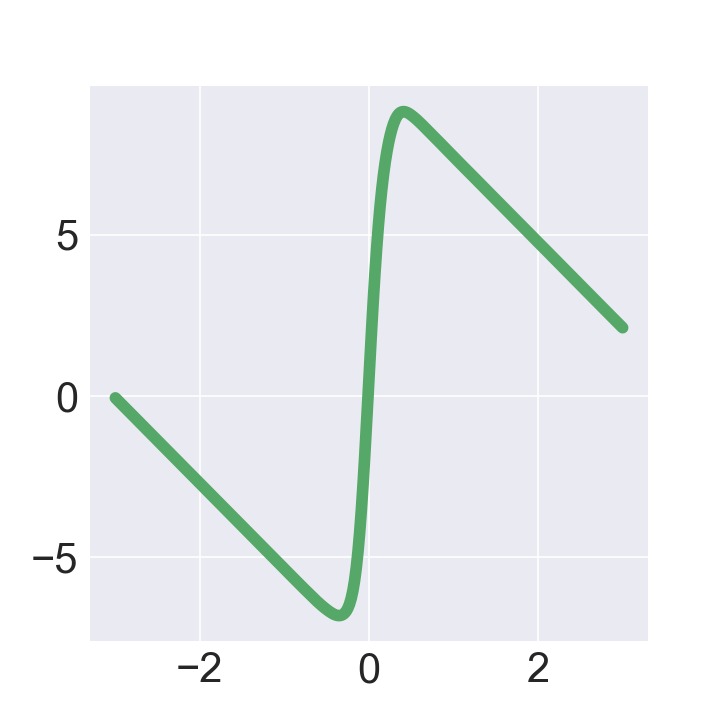}
    \includegraphics[width=0.24\textwidth]{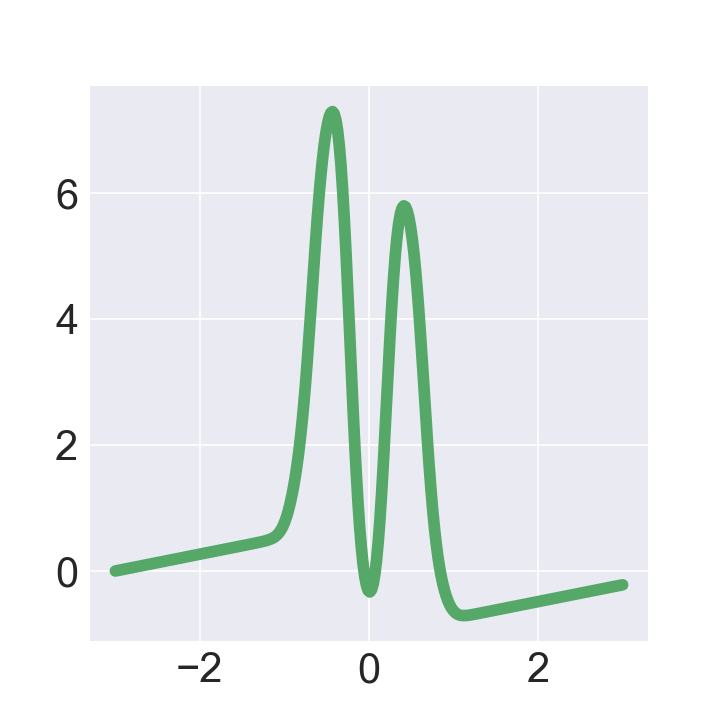}
    \includegraphics[width=0.24\textwidth]{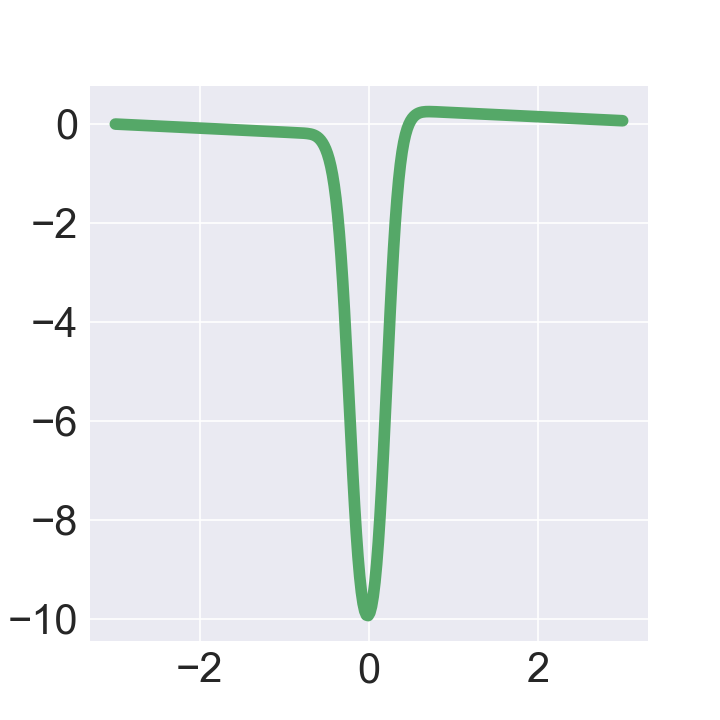}
    \includegraphics[width=0.24\textwidth]{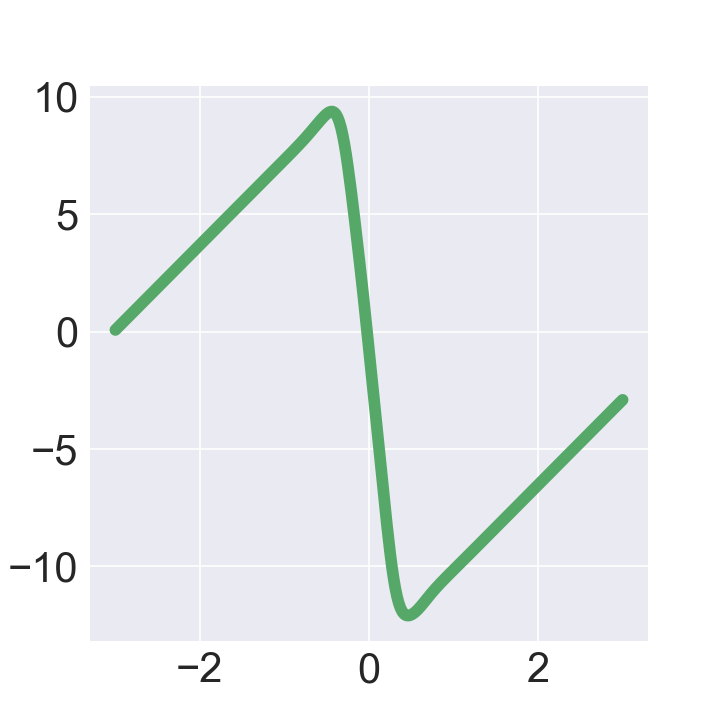}
    \caption{Activation/Potential Function}
    \label{sfig:actfun}
  \end{subfigure}
  \caption{Visualization of filters (left) and activation functions (right) of our model. Filters: Top to bottom: Filters of the disparity map, filters of the RGB color image and filters of the confidence map. Activation Functions: Visualization learned activation functions. The first row (blue) shows the activation functions in the derivative space. The second row (green) shows the integrate of the same functions which correspond to the potential functions in the energy domain. The structure is even better visible if zoom is decreased in a pdf viewer.}
  \label{fig:filters}
\end{figure}

We give further insights into what and how our model learns by visualizing the filters and activation functions learned by our model.
\cref{sfig:filters} shows selected filter kernels. 
The first row contains filters for the disparity map, the second row contains filters for the RGB color image and the third row contains filters for the confidence map.
Note that the learned filters are interpretable because they contain structure.
This suggests that our model captures statistics of how to appropriately denoise disparity maps, confidence maps and color images jointly.

\cref{sfig:actfun} shows the learned activation functions. 
We can integrate the learned activation functions (blue) to get the potential functions (green) used in our energy. 
For example the third potential function has the shape of a truncated Huber function. 
This means the function has learned to be robust against outliers.

\section{Additional Qualitative Results}
\begin{figure}[ht]
  \centering
  \includegraphics[width=0.32\textwidth]{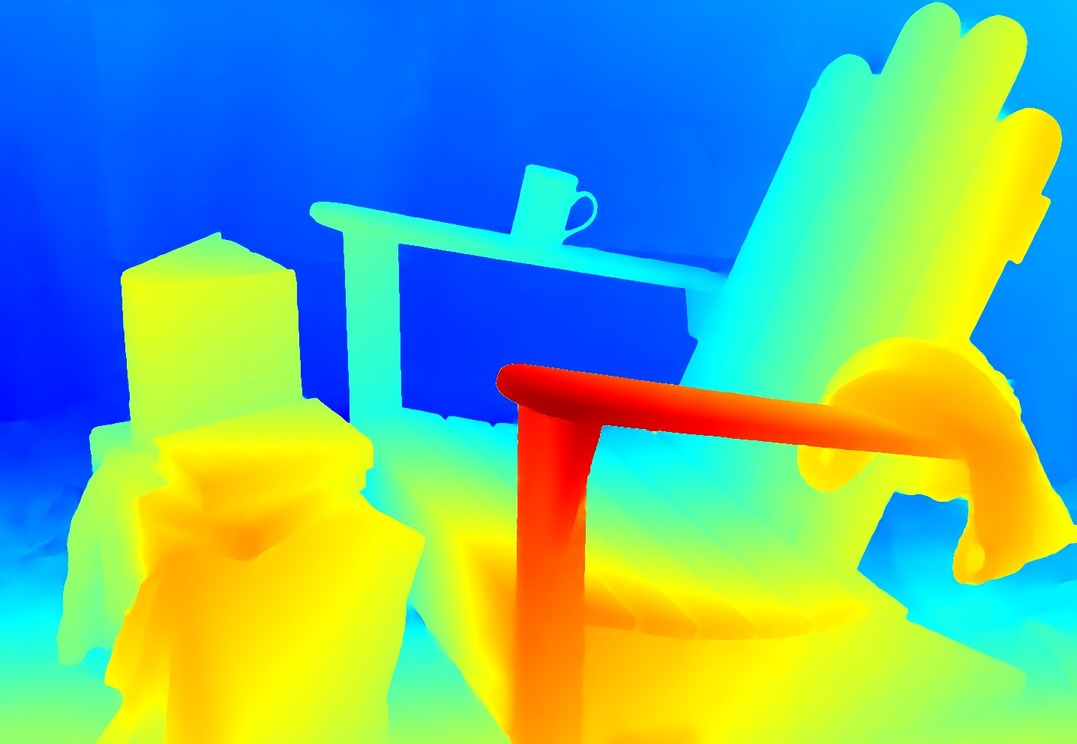}
  \includegraphics[width=0.32\textwidth]{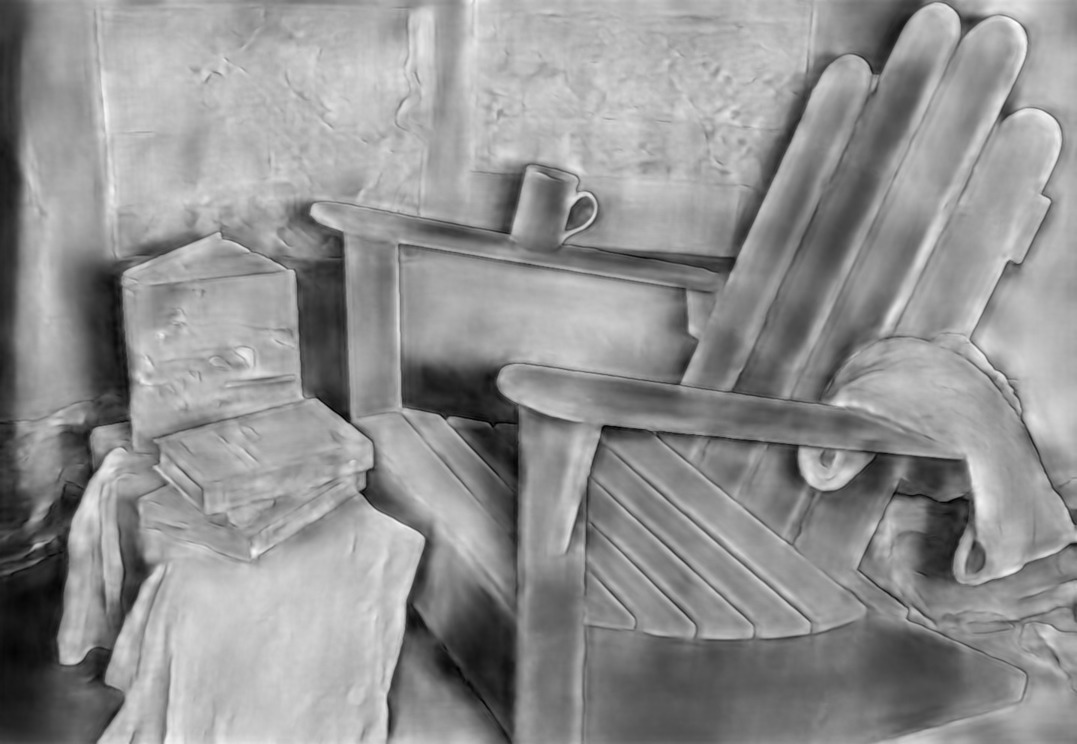}
  \includegraphics[width=0.32\textwidth]{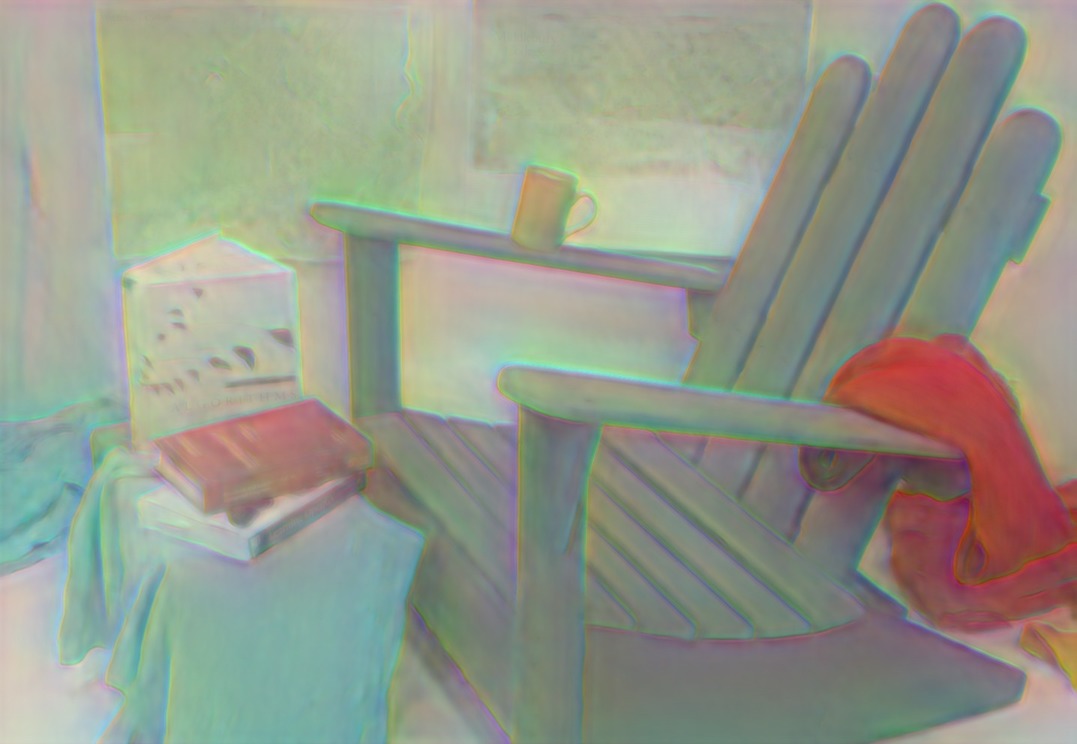}

  \includegraphics[width=0.32\textwidth]{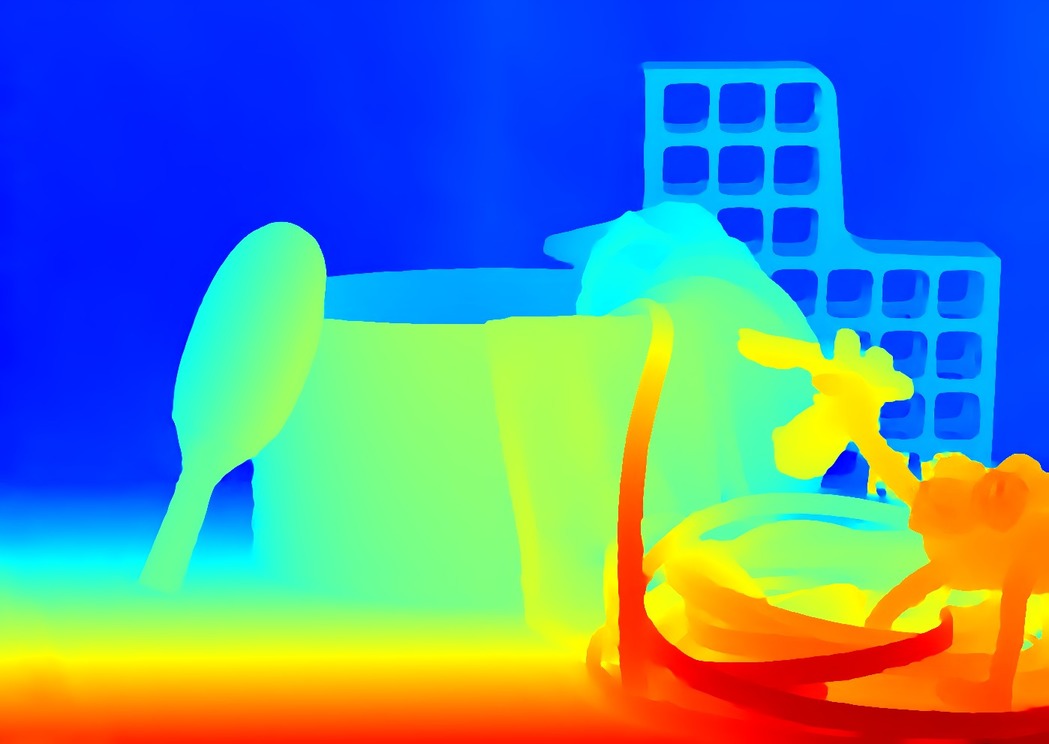}
  \includegraphics[width=0.32\textwidth]{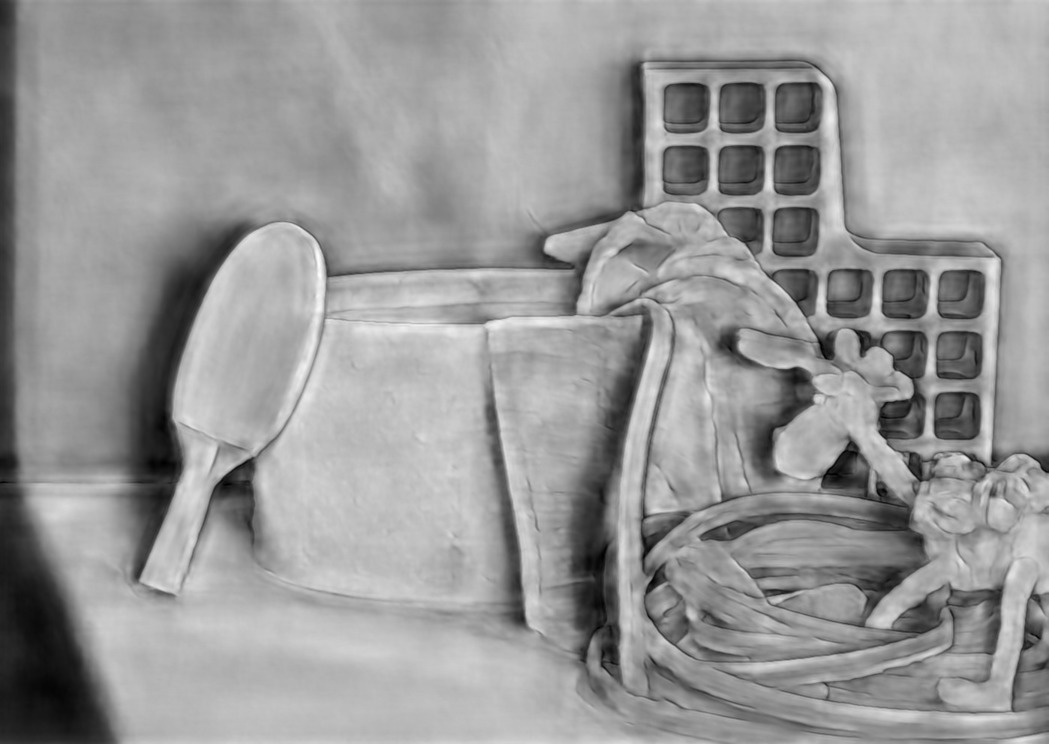}
  \includegraphics[width=0.32\textwidth]{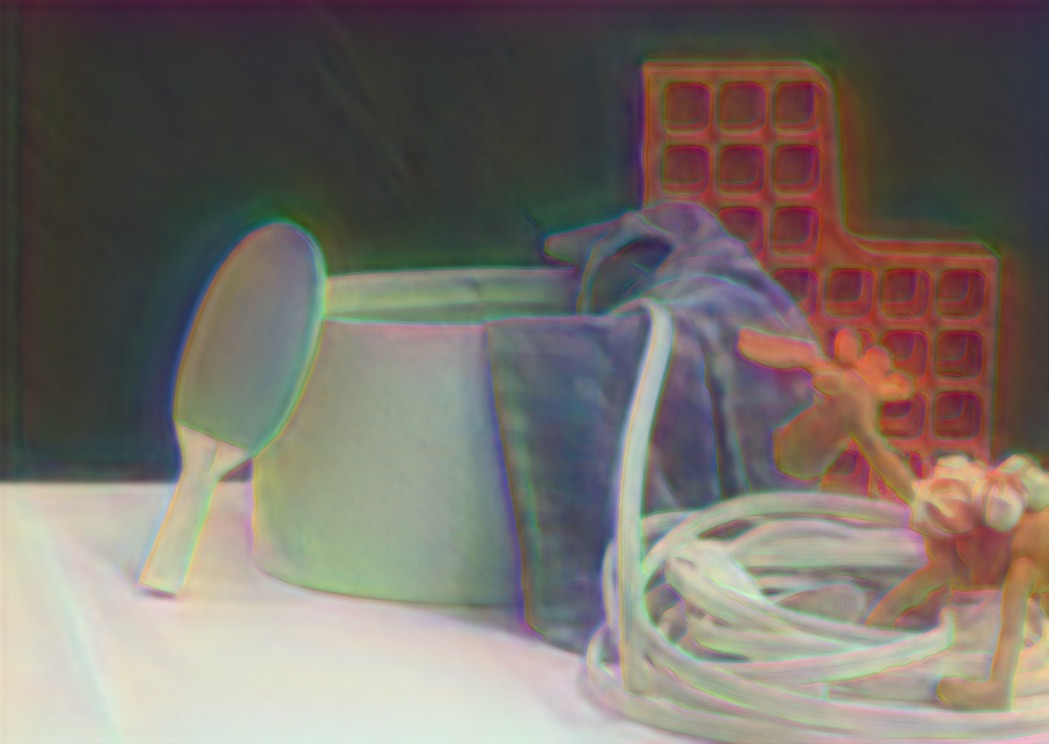}

  \includegraphics[width=0.32\textwidth]{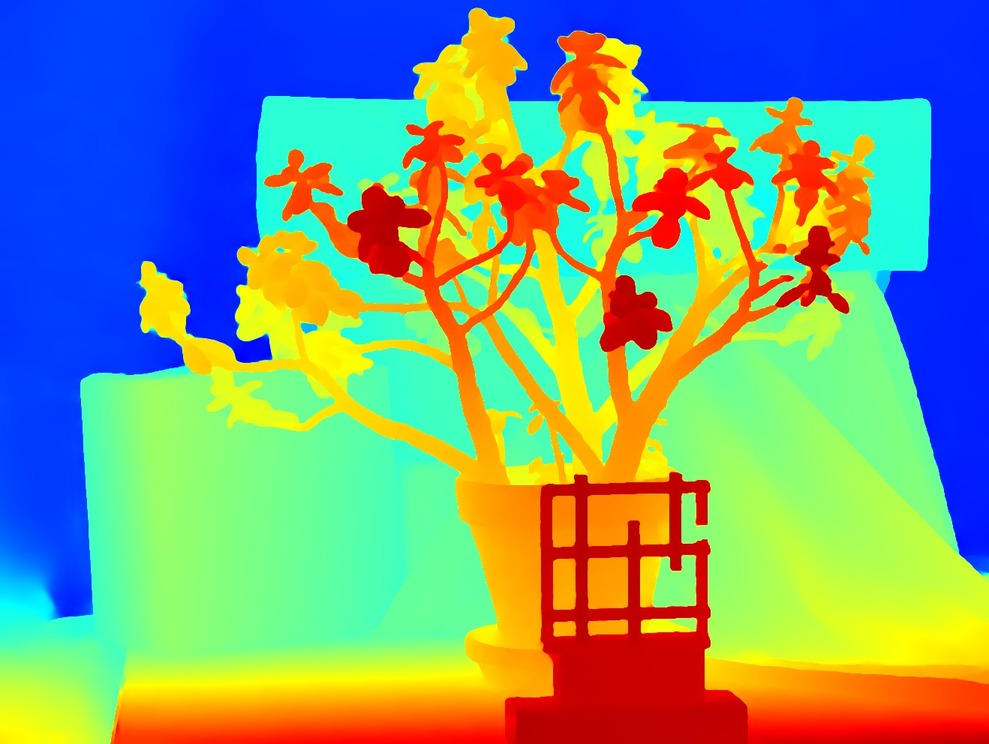}
  \includegraphics[width=0.32\textwidth]{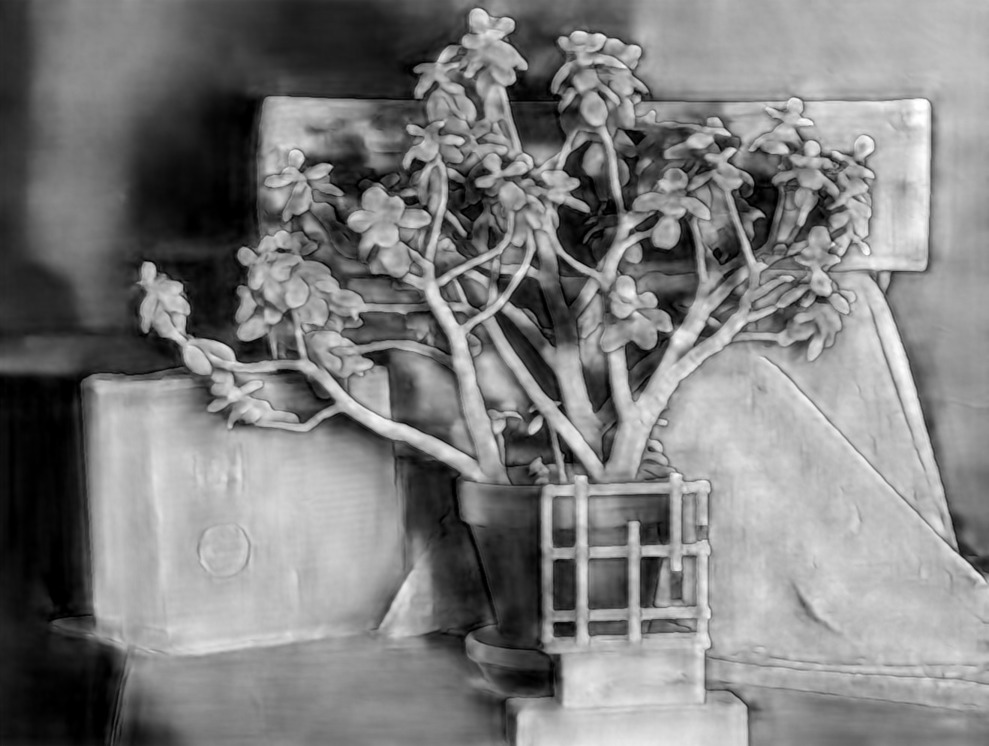}
  \includegraphics[width=0.32\textwidth]{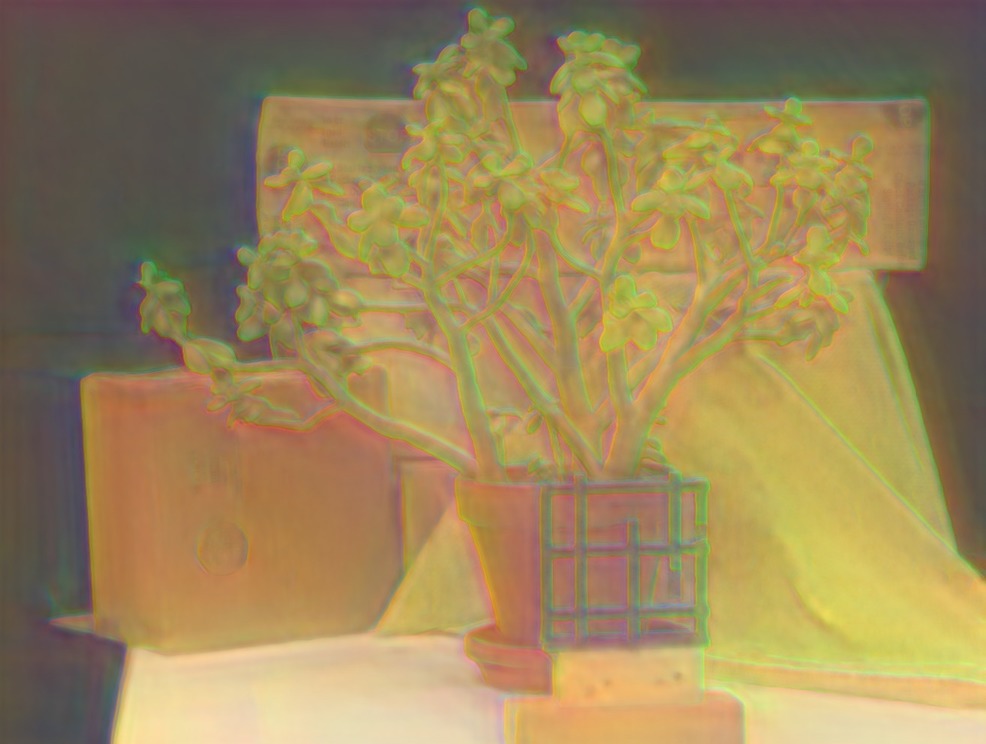}

  \includegraphics[width=0.32\textwidth]{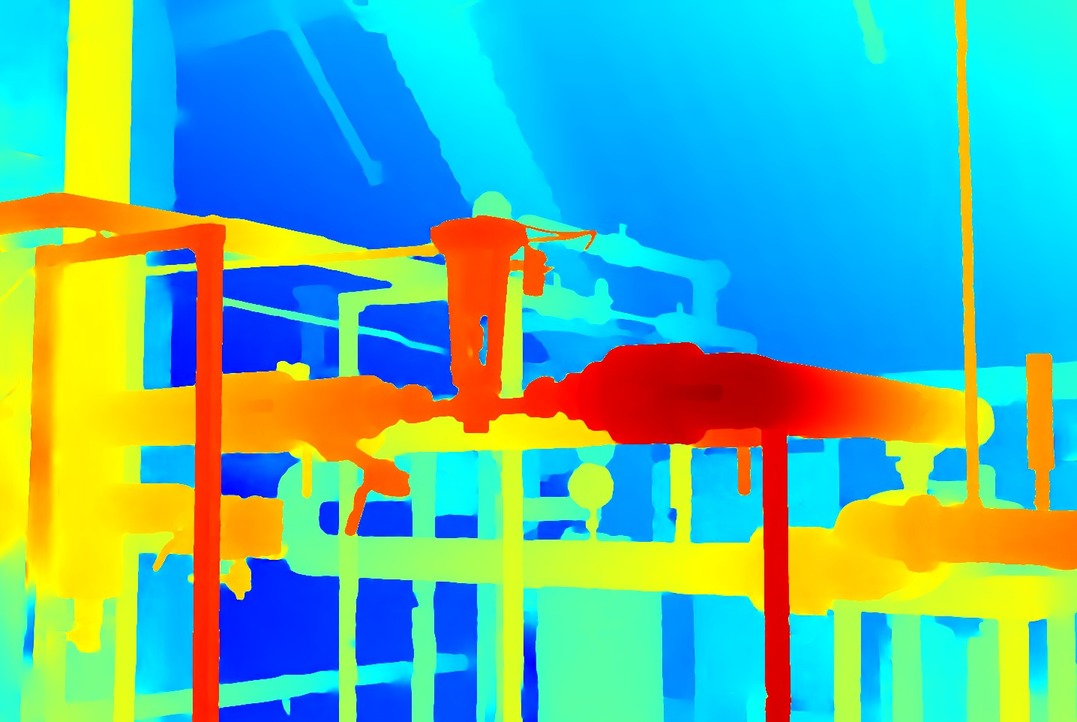}
  \includegraphics[width=0.32\textwidth]{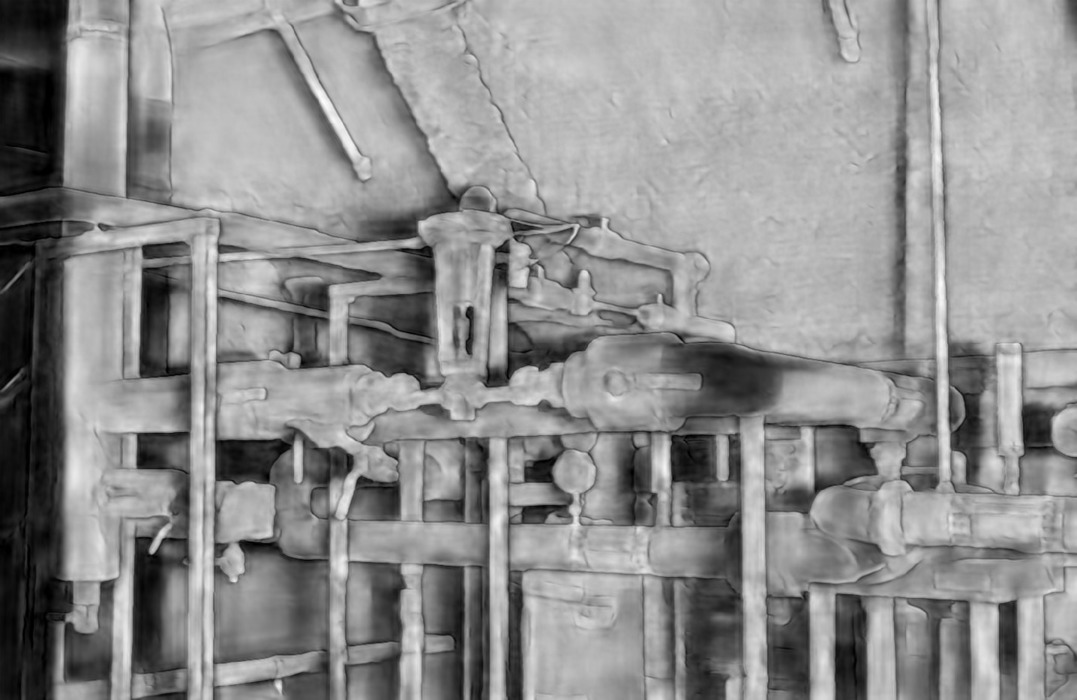}
  \includegraphics[width=0.32\textwidth]{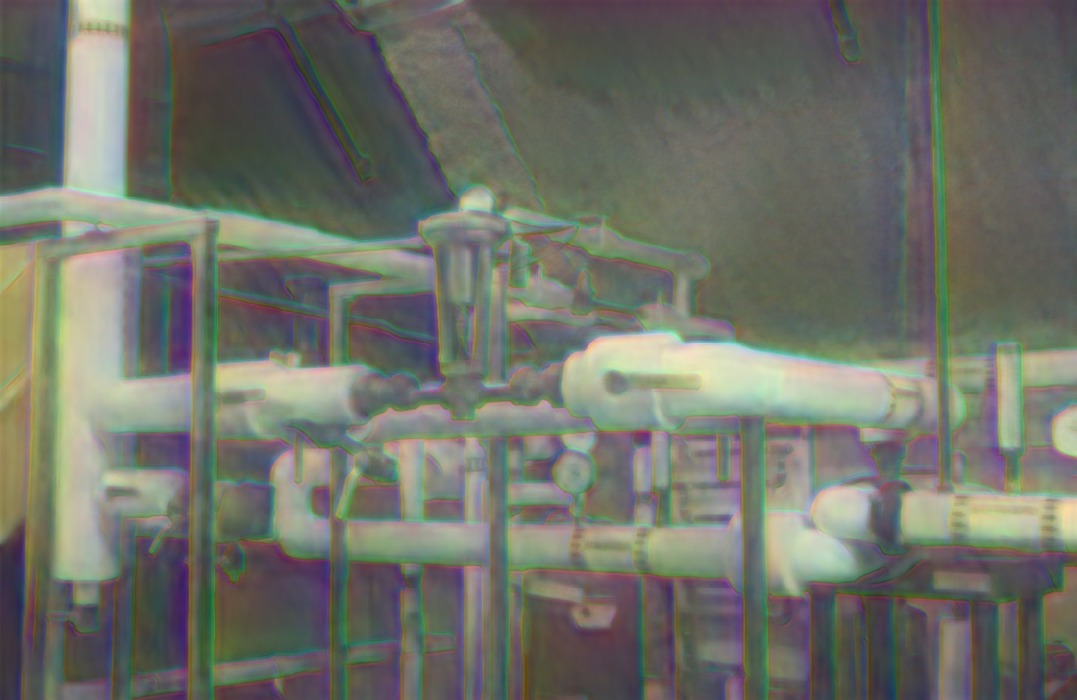}

  \includegraphics[width=0.32\textwidth]{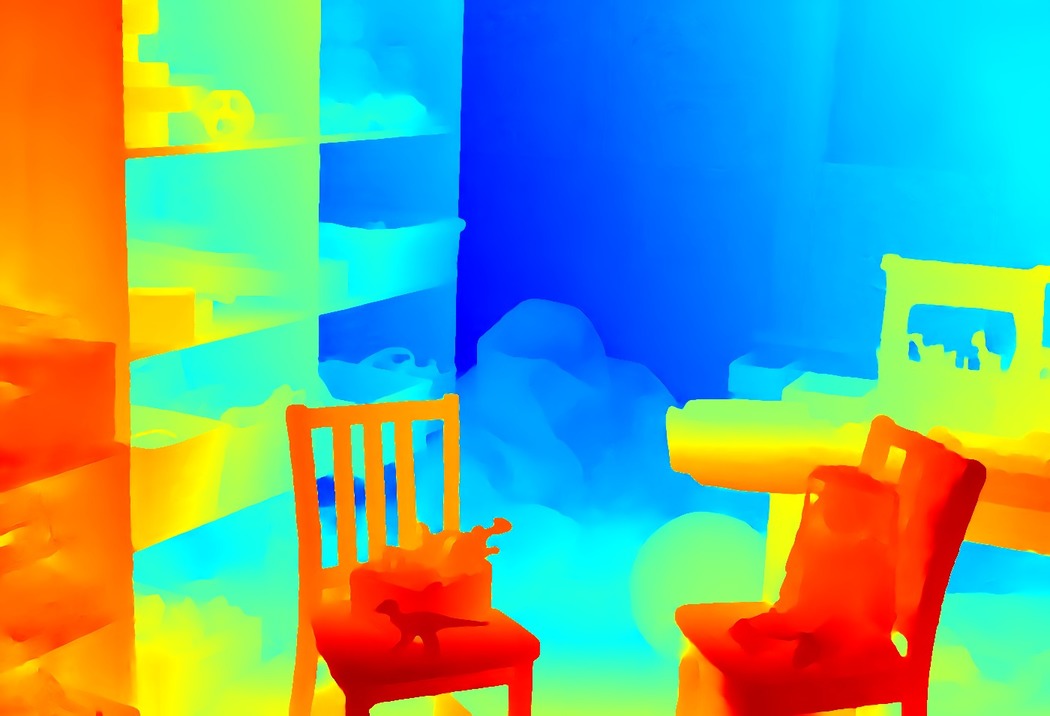}
  \includegraphics[width=0.32\textwidth]{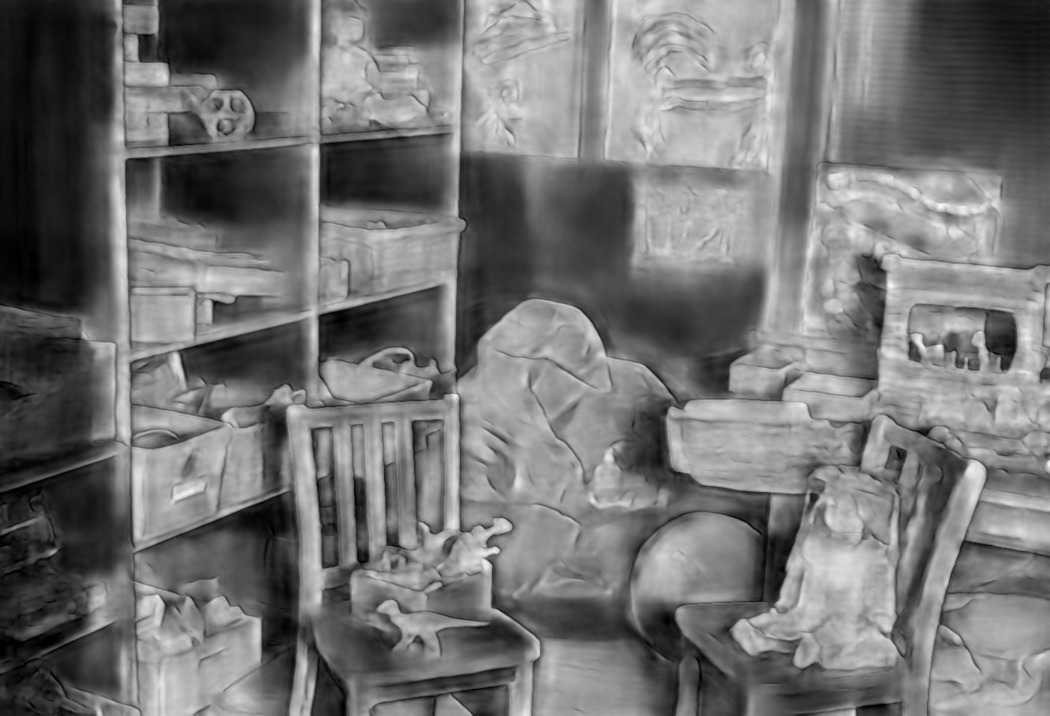}
  \includegraphics[width=0.32\textwidth]{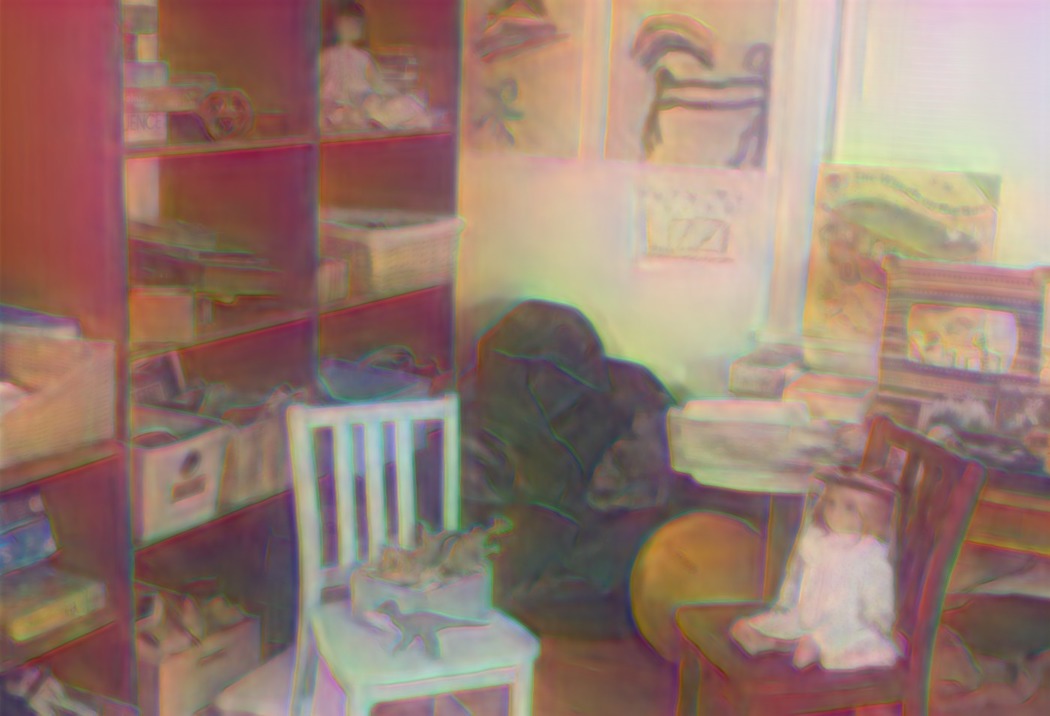}

  \includegraphics[width=0.32\textwidth]{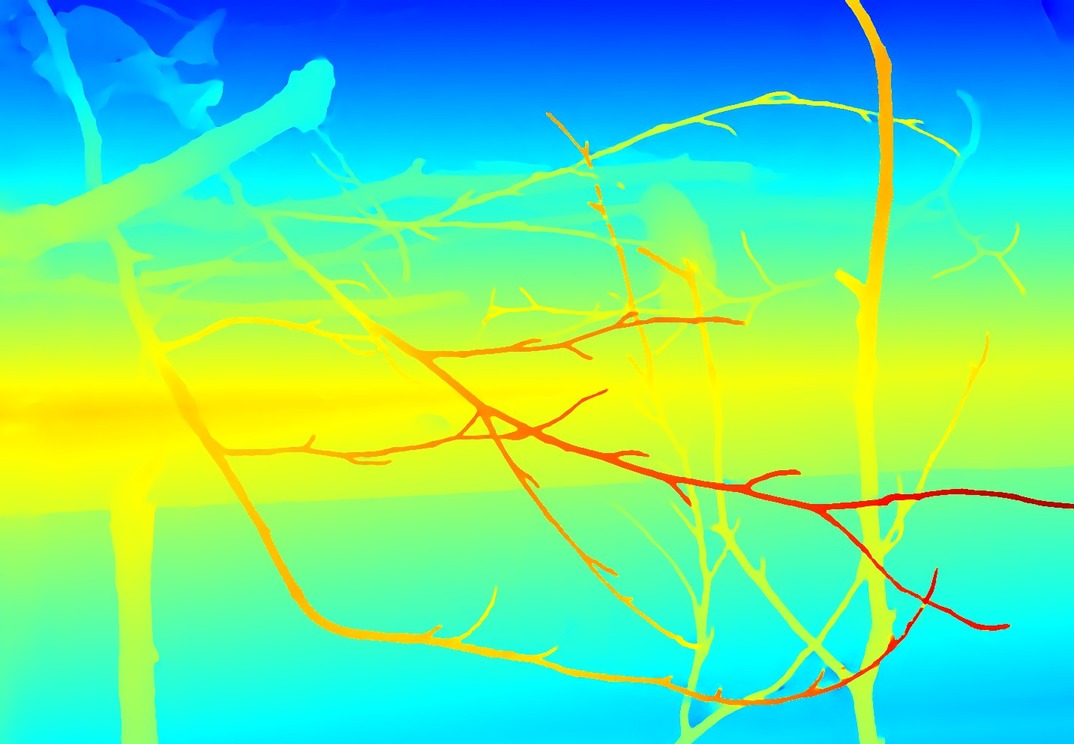}
  \includegraphics[width=0.32\textwidth]{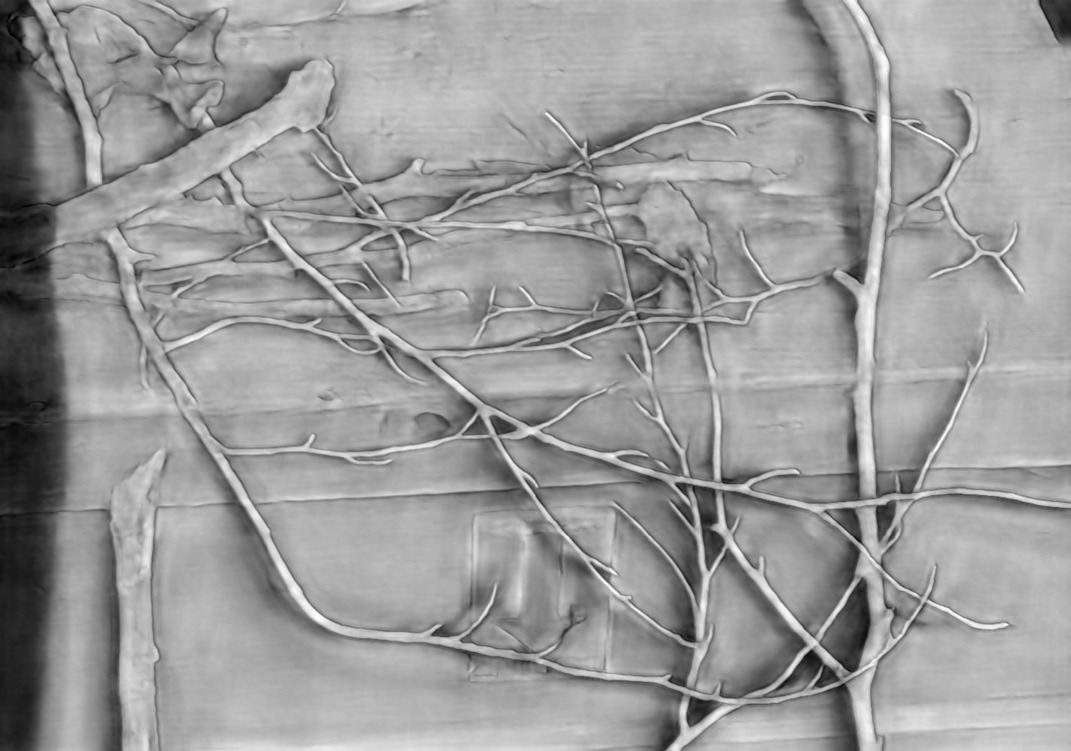}
  \includegraphics[width=0.32\textwidth]{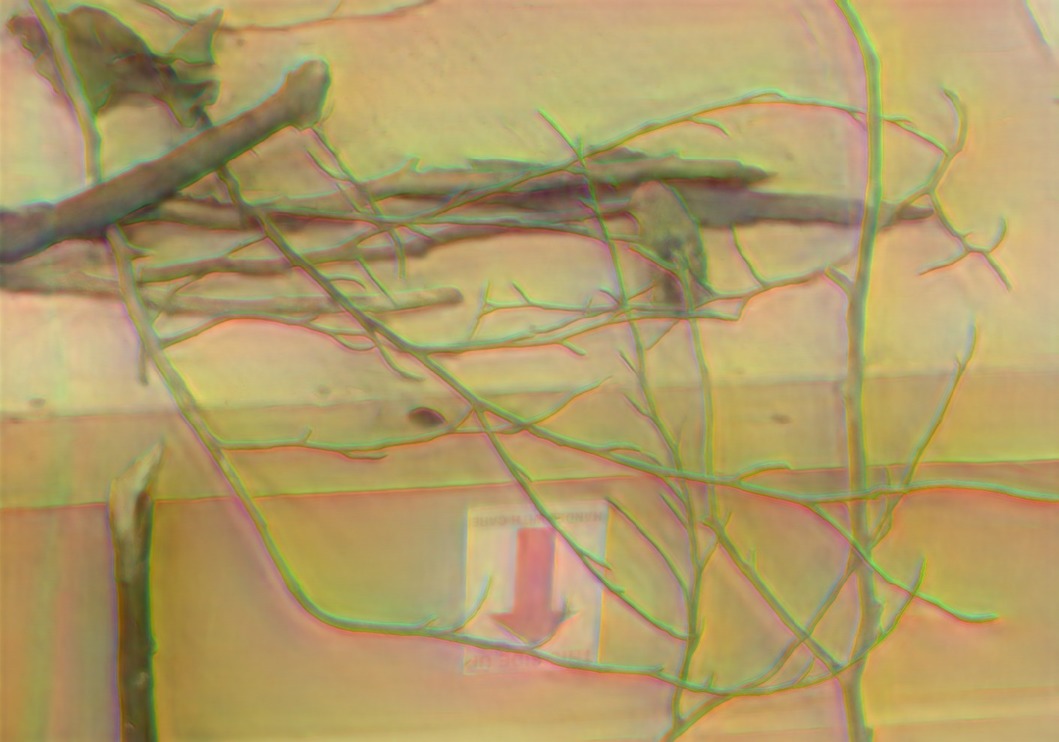}

  \caption{Results of VN$^{7,11}_4$ on half size (H) Middlebury images. Left to right: Disparity map, confidences and color image. Our model learns to use object edges to guide the denoising of the disparity map. Best viewed with zoom on the PC.}
  \label{fig:sup:vnvisMB}
\end{figure}

\begin{figure}[ht]
  \centering
  \includegraphics[width=0.4\textwidth, clip, trim={700 16 16 16}]{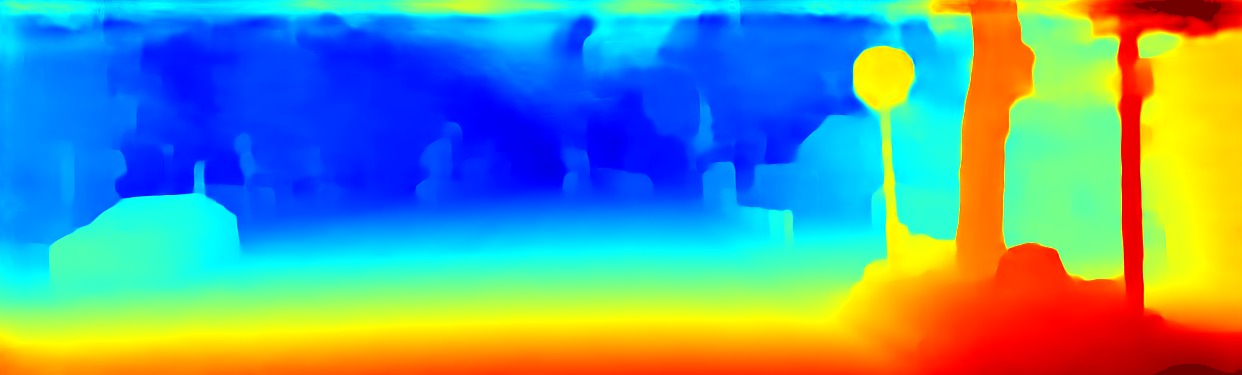}
  \includegraphics[width=0.4\textwidth, clip, trim={700 24 16 16}]{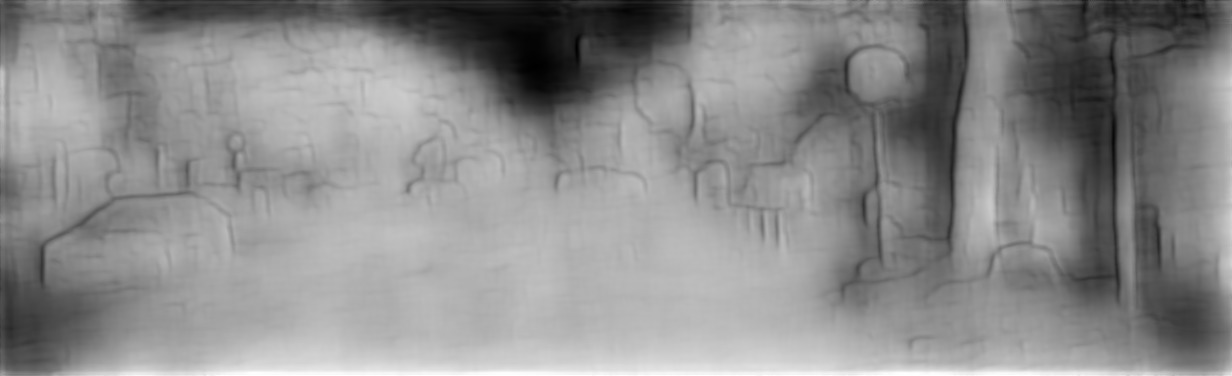}
  \includegraphics[width=0.4\textwidth, clip, trim={700 16 16 16}]{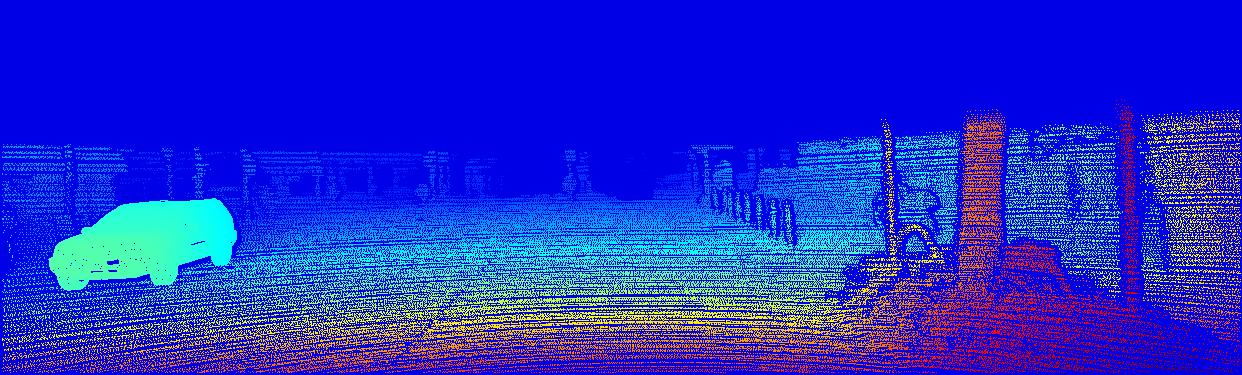}
  \includegraphics[width=0.4\textwidth, clip, trim={700 24 16 16}]{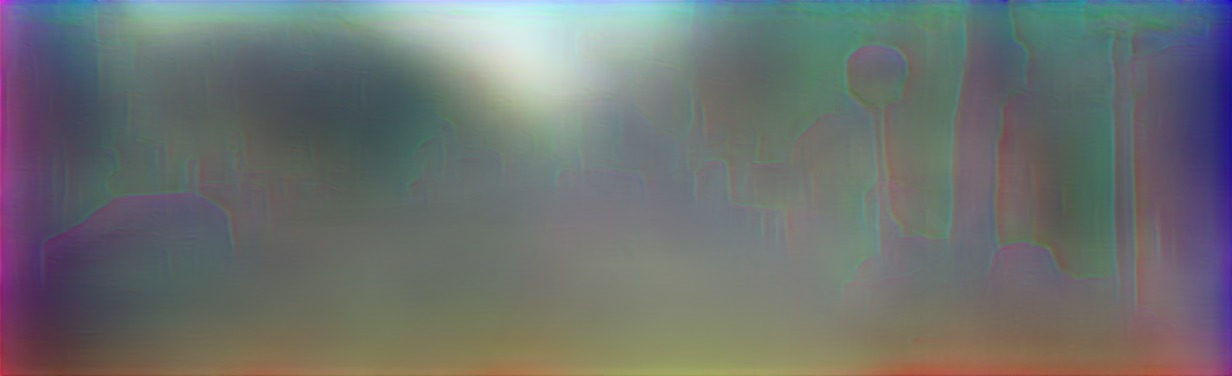}

  \vspace{0.5em}
  \includegraphics[width=0.4\textwidth, clip, trim={16 16 700 16}]{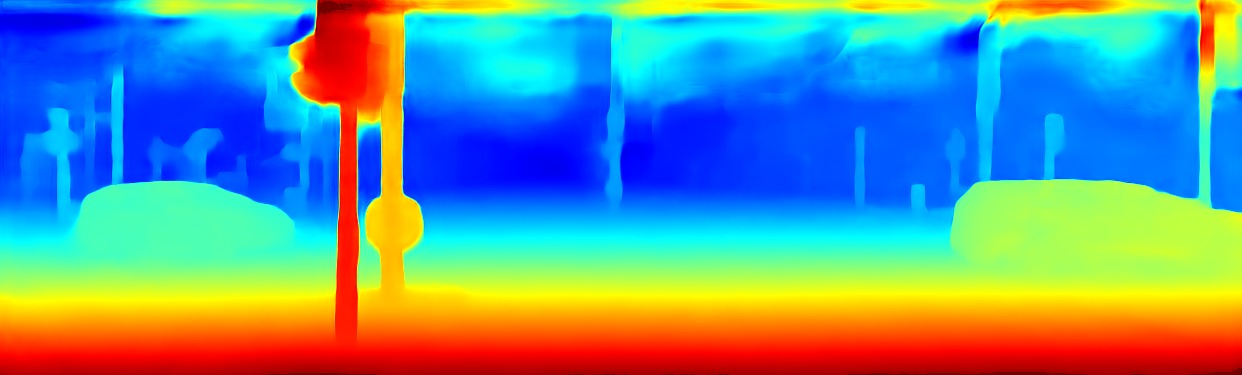}
  \includegraphics[width=0.4\textwidth, clip, trim={16 24 700 16}]{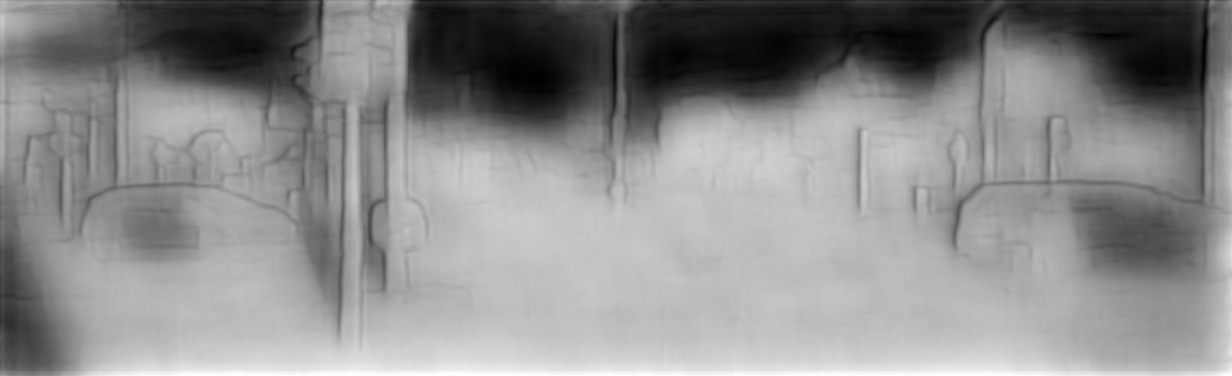}
  \includegraphics[width=0.4\textwidth, clip, trim={16 16 700 16}]{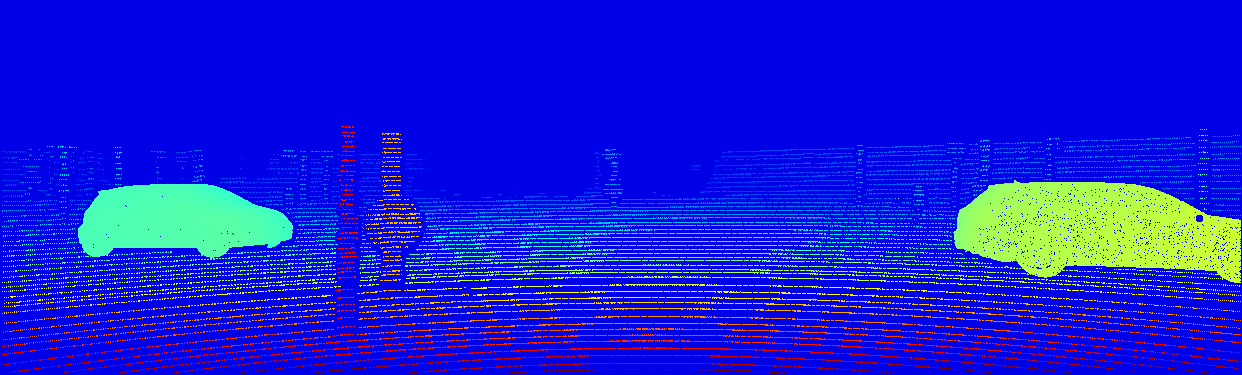}
  \includegraphics[width=0.4\textwidth, clip, trim={16 24 700 16}]{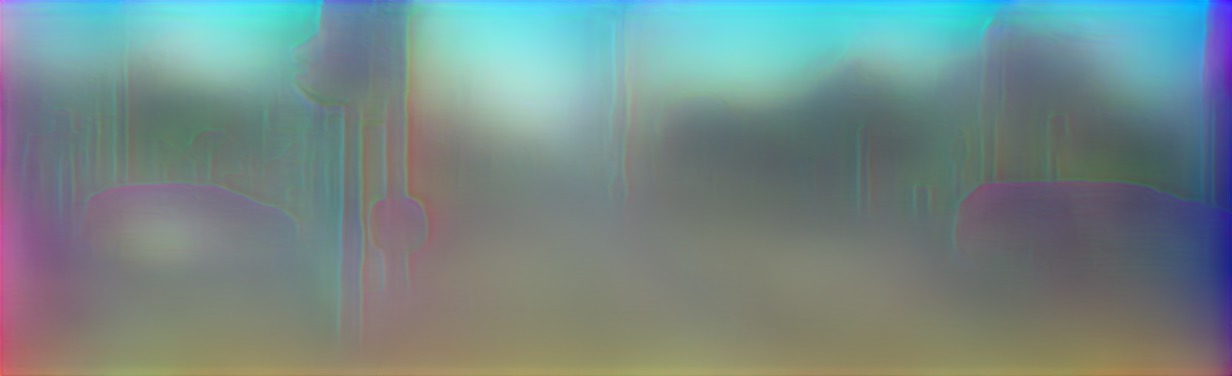}

  \vspace{0.5em}
  \includegraphics[width=0.4\textwidth, clip, trim={700 16 16 16}]{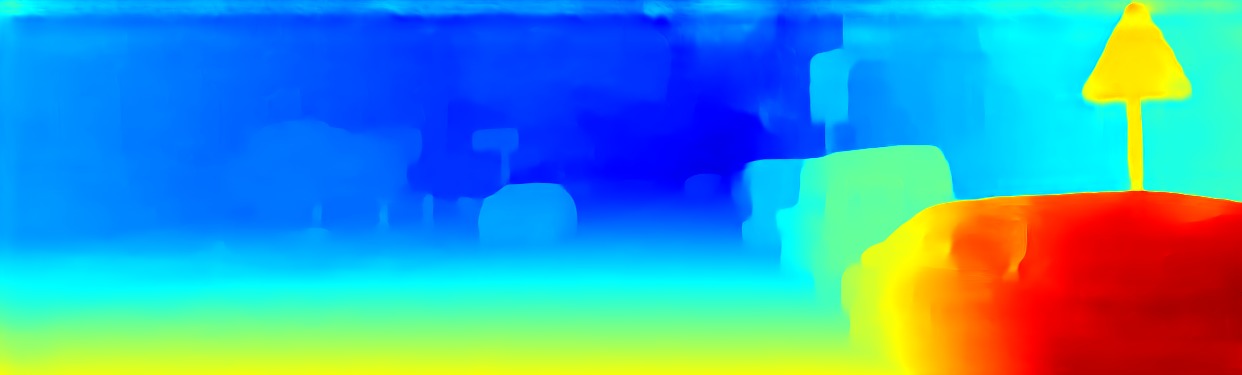}
  \includegraphics[width=0.4\textwidth, clip, trim={700 24 16 16}]{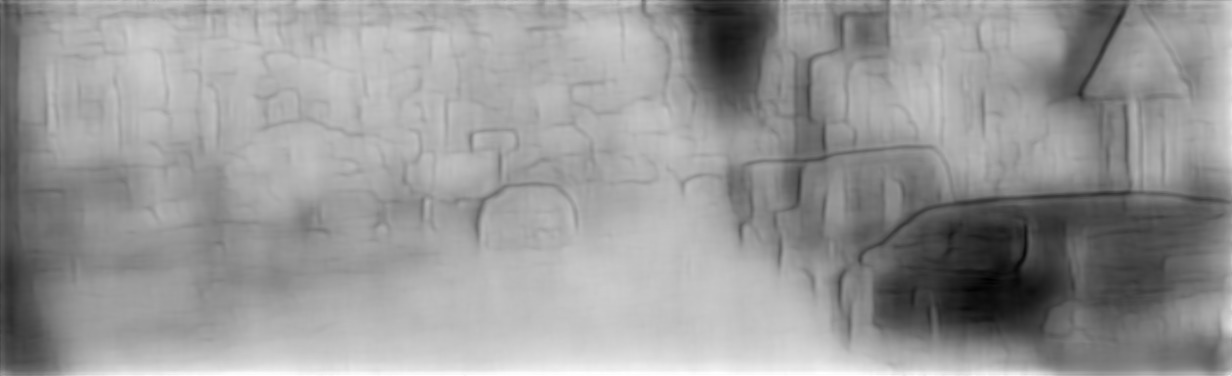}
  \includegraphics[width=0.4\textwidth, clip, trim={700 16 16 16}]{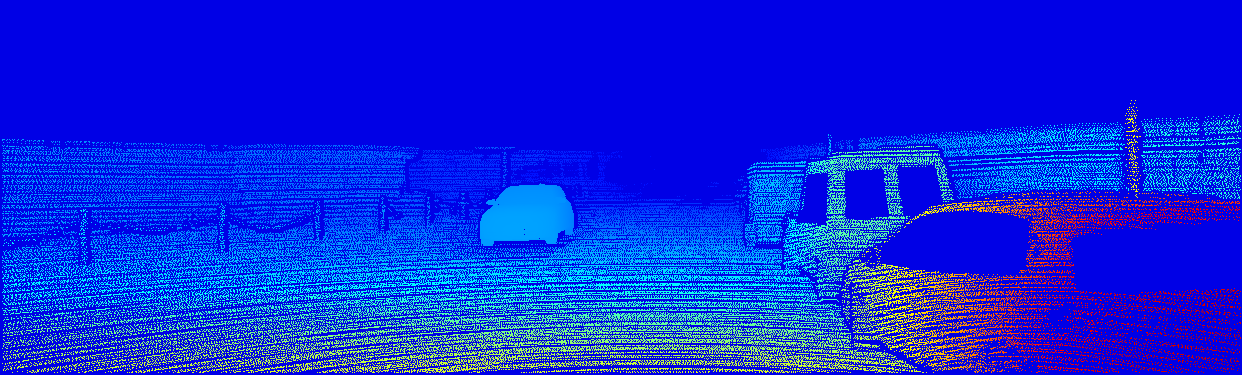}
  \includegraphics[width=0.4\textwidth, clip, trim={700 24 16 16}]{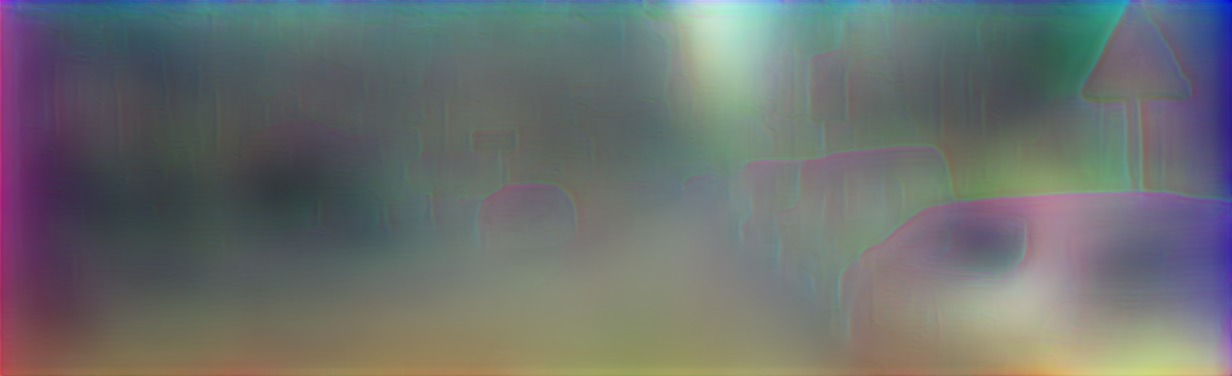}
\caption{Results of VN$^{7,5}_4$ on the Kitti 2015 dataset. Starting at top-left in clock-wise order: VN disparity maps, confidence map, color images, ground-truth. Note how strong object boundaries are detected in the confidence image and in the RGB image to guide the disparity refinement. Best viewed on the PC.}
\label{fig:sup:KittiVis}
\end{figure}

We use the supplementary material to show additional qualitative results of our method.
The results shown in \cref{fig:sup:KittiVis,fig:sup:vnvisMB} impressively show how our simple denoising/refinement scheme is able to reconstruct high quality disparity maps.

The results on the Middlebury benchmark show, how information is exploited from both, the confidences and the input image to reconstruct the best possible disparity map. 
Note how the blue channel in \cref{fig:sup:vnvisMB} seems to be an indicator for regions being occluded, because they appear always next to left sided object boundaries.

As shown in \cref{fig:sup:KittiVis} the results on Kitti show similar properties. 
Note here how fine details in the background are accurately captured. 
The confidence maps and the images guide the disparity map where discontinuities are likely to occur and where smoothing is the better option.
The former is always close to object boundaries, and the latter is on the street, on cars or in the background.

\end{document}